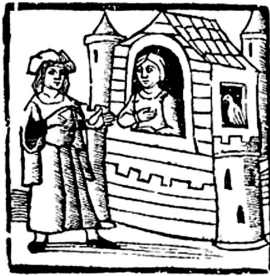

# The Life of *Lazarillo de Tormes* and of His Machine Learning Adversities

## Non-traditional authorship attribution techniques in the context of the *Lazarillo*

Javier de la Rosa & Juan Luis Suárez
The University of Western Ontario, London, Canadá

RESUMEN:

Obra cumbre del Siglo de Oro español y precursora de la así llamada novela picaresca, *La vida de Lazarillo de Tormes y de sus fortunas y adversidades* aún continúa como novela anónima. Multitud de investigadores le han atribuido distintos autores basándose en una plétora de criterios, sin embargo, no se ha conseguido alzanzar un consenso. La lista de posibles candidatos se ha ido nutriendo a lo largo del tiempo, aunque no todos cuenten con el mismo apoyo por parte de la comunidad investigadora. En este estudio partimos del conocimiento de los expertos en la materia para constituir un grupo de posibles candidatos cuyas obras son estudiadas desde un punto de vista computacional. El análisis de los textos aplicando técnicas de aprendizaje automático de marcas estilísticas y estilografía permite arrojar algo de luz sobre la autoría del *Lazarillo*. Los distintos métodos son a su vez analizados y sometidos a discusión de acuerdo al rendimiento que proporcionan en nuestro caso específico. De acuerdo a nuestra metodología, el autor más probable parece ser Juan Arce de Otálora, seguido muy de cerca por Alfonso de Valdés. Desafortunadamente, el método establece que no se puede dar una atribución certera con el corpus dado.

PALABRAS CLAVE: *Lazarillo de Tormes*, atribución de autoría, estilografía, aprendizaje automático.

ABSTRACT:

Summit work of the Spanish Golden Age and forefather of the so-called picaresque novel, *The Life of Lazarillo de Tormes and of His Fortunes and Adversities* still remains an anonymous text. Although distinguished scholars have tried to attribute it to different authors based on a variety of criteria, a consensus has yet to be reached. The list of candidates is long and not all of them enjoy the same support within the scholarly community. Analyzing their works from a data-driven perspective and applying machine learning techniques for style and text fingerprinting, we shed light on the authorship of the *Lazarillo*. As in a state-of-the-art survey, we discuss the methods used and how they perform in our specific case. According to our methodology, the most likely author seems to be Juan Arce de Otálora, closely followed by Alfonso de Valdés. The method states that not certain attribution can be made with the given corpus.

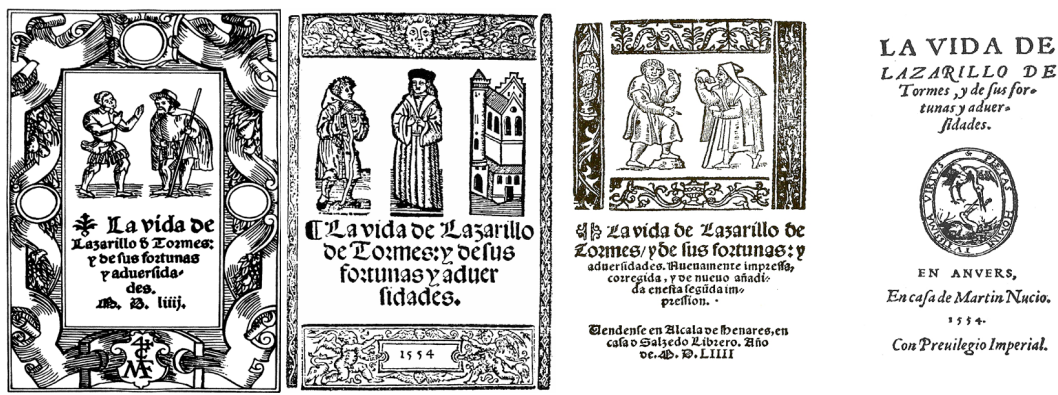

> *It would be much better to build up results*
> *one centimetre at a time on a base one kilometre wide,*
> *than to build up a kilometre of research on a one-centimetre base.*
> — Roberto Busa, 1980

## Introduction

The authorship of *The Life of Lazarillo de Tormes and of His Fortunes and Adversities* —usually referred to as the *Lazarillo de Tormes*, or just (and henceforth) the *Lazarillo*— is a topic that has interested researchers ever since the story was first published. The earliest preserved editions were printed in 1554 in Burgos (Spain), Alcalá de Henares (Spain), Medina del Campo (Spain),[1] and Antwerp (Belgium), although there might be at least two earlier editions yet to be found that complete the phylogenetic tree (figure 1 shows a possible stemma).[2] After a short period of popularity, in 1559 it was added to the *Index* of forbidden books compiled by the Inquisitor General Fernando de Valdés,[3] and therefore banned from public circulation due to its acid anti-clerical criticism.[4] The text's religious aspects have been particularly influential in scholars' attempts to create an accurate profile of the anonymous writer. The author has been therefore considered to be a converted

1.– The edition of Medina del Campo is the newest found. It appeared in 1992 hidden inside a wall, together with other texts that could be considered problematic by the Inquisition (Cañas Murillo).

2.– It is believed that the editions of 1554 are actually second editions following the success of a first edition of the book published as early as 1538, as suggested by Navarro Durán as the post quem of the little book: «el autor sólo puede referirse a las primeras [Cortes] porque no sabe que se van a celebrar unas segundas, ya que el Lazarillo se escribió antes de 1538,» («the author can only be referring to the the first [Cortes] as he does not know that there will be second ones, due to the fact that the Lazarillo was written before 1538») (Navarro Durán 2002a, 7-13). See also the analysis by Francisco Rico in his introduction to his edition (Anónimo ed. Rico, 13-15), or the section «Las ediciones desconocidas» by José Caso González (Anónimo ed. Caso González, 11-14; Caso González, «La primera edición» 189-206). More recently, Arturo Rodríguez and Alfredo Rodríguez López-Vázquez based on weak documental proof (not the edition itself) and stemmatics supported an earliest edition in 1550 (Rodríguez and Rodríguez López-Vázquez).

3.– Later Rome's *Index Librorum Prohibitorum* by Pope Pius VI also included books that could be re-edited prior partial censorship.

4.– See for example Manuel J. Asensio («La intención» 78-102) and Víctor García de la Concha (243-77). Reyes Coll-Tellechea argues that the real reason for the addition of the *Lazarillo* to the *Index* was the publication of the second part *Segunda Parte del Lazarillo de Tormes*, which was read as a political provocation and therefore never released again until the end of the *Index* («The Spanish» 75-97).

Jew (Castro, «Perspectiva» 123-138; «Hacia Cervantes» 149-166), an illuminist (Asensio, «La intención religiosa» 78-102; Asensio, «Más sobre el Lazarillo» 245-50), or an erasmist (Márquez Villanueva, 107-I37), but these theses have been deeply questioned by acclaimed critics such as Marcel Bataillon and Eugenio Asensio, who depict the author as a humanist (Bataillon, «Erasmo y España» 609-611; «Novedad y fecundidad» 1-25; Pícaros y picaresca 215-243; Asensio, «El erasmismo» 31-99; Asensio, «La peculiaridad» 339-343). Nevertheless, the notion of an author in contact with such spiritual and ideological inſterests still persists in the literature, which could have informed their decision not to sign the little book.

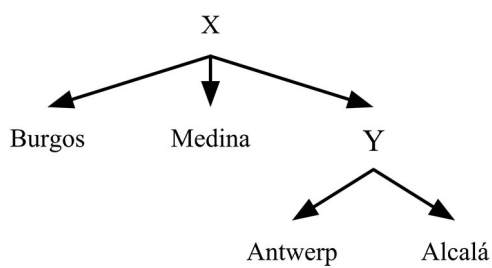

Figure 1: Stemma for the editions of the *Lazarillo* of 1554 as structured by Jesús Cañas Murillo. *X* and *Y* denote lost editions, being *X* the *editio princeps* or «true first edition».[5] Aldo Ruffinatto's stemma also takes into consideration Juan López de Velasco's *Lazarillo Castigado* after an analysis following the principles of *ecdótica* (ecdotic analysis) (Anónimo ed. Aldo Ruffinatto; Ruffinatto, «La princeps» 249-96; «Algo más» 523-36).

It was not until 1573 that a censored version was circulated again in Spain, but omitting treatises 4 and 5 and assorted paragraphs from other parts of the book. Juan López de Velasco, Philip II of Spain's cosmographer and historian, was the person responsible for the trimming of the *Lazarillo*, whose edition is known as the *Lazarillo Castigado* (*Lazarillo Punished*) (Asensio, «La intención»). The exerted censorship was subtle but radical as it transformed the identity of the *Lazarillo* turning the protagonist from «a victim of the socio-economic circumstances into a Lázaro responsible of his own social and moral condition» («[Dichas alteraciones] estaban dirigidas a transformar la imagen de un Lázaro víctima de las circunstancias socioeconómicas en un Lázaro responsable de su condición

---

5.– In Cañas Murillo (134):

El texto de Medina del Campo no procede directamente de ninguna de las versiones hasta ahora conservadas. Forma una rama textual independiente. Dada su proximidad a Burgos, que, procede directamente del arquetipo X perdido, y la mayor limpieza de sus lecciones, parte de las cuales coinciden significativamente con Amberes, más corregido, insistimos, que Burgos y Alcalá, hay que concluir que dicha rama hay que hacerla depender también directamente del arquetipo X.

(The text of Medina del Campo does not come from any of the versions preserved until now. It constitutes an independent textual branch. Given its proximity to that of Burgos, which comes from the lost archetype X, and the greater cleanness of its lessons, many of which significantly match with Amberes, more proofread, we insist, than Burgos and Alcalá, we conclude that such branch must depend on the archetype X too.)

social y moral.»)[6] Although the work by Juan López de Velasco allowed the *Lazarillo* to leave the list of forbidden books, by then the Antwerp's edition, translated to different languages, had already spread over Europe.[7] It is suggested that the book that actually started the picaresque novel and influenced so many authors afterwards was in fact the censored edition. Until the final abolition of the Inquisition and the end of the *Index* in 1834, the *Lazarillo Castigado* was the only edition officially available in Spain for more than 250 years. If the *Lazarillo Castigado* was indeed the seed of the picaresque genre, then we would possibly have a preliminary explanation for two gaps unaccounted for: first, the time elapsed between the publication of *Lazarillo* in 1554 and the appearance in 1626 of the next considered picaresque novel, *The Swindler* (*El Buscón*) by Quevedo; and second, the difference between the deterministic style of the *Lazarillo* and the cruel reality that punishes the rogue for his aspirations in the following titles that became later on a more common topic in the genre. Unfortunately, the argument of *Lazarillo Castigado* being the book that started the picaresque genre does not count with the discussion around the date of the *princeps* and relies heavily on the idea of nobody using alternative channels of distribution or being rebellious against the establishment. Given the circumstances involved in the discovery of the edition of Medina del Campo we must take this suggestion with uncertainty (Alberto Martino, *Lazarillo*). Nevertheless, the importance of the figure of Juan López de Velasco does not end with his cleverly expurgated edition, as we will see soon.

## A Book by Many Authors

The list of possible authors has grown with the years along with the painstaking effort of many researchers who devoted their time, intelligence, and expertise —sometimes even through their entire careers (see tables 1 and S1)— to this text. A noble and scientific goal has guided them to put an end to the enigma and to unveil the true identity of the author of the *Lazarillo*. These 400 years of attributions have left us an insane, nearly intractable, amount of bibliography that must be reviewed and studied before dreaming of making a contribution to the state-of-the-art. It has become very hard to avoid certain feelings of genuine *argumentum ad verecundiam*, at least in the initial stages of the research. This amount of bibliography, paradoxically, instead of keeping novel scholars away has not been a deterrent and new proposals are still being added to the pool of candidates today, although some of them using modern and less explored methods (mostly computational) that were not available a decade or so ago. It is with respect to these techniques that we try to limit ourselves in the present study.

Chronologically, the first attribution occurred more than half a century after the earliest known edition. In 1605 the Hieronymite Friar José de Sigüenza was the first to propose a possible author: the also friar, Juan de Ortega. Father Sigüenza's *Historia de la Orden*

---

6.– See Coll-Tellechea («Lazarillo Castigado» 32-33). Others limit the extent to which the trimming affected the story of Lázaro (Agulló y Cobo, *A vueltas*).

7.– By 1596 there were already editions published in London (England) with Diego Hurtado de Mendoza as the author. See chapter 2 of the precise and exhaustive work by Alberto Martino, and also his second volume dedicated to the reception of the *Lazarillo* in Europe.

*de San Jerónimo* (*History of the Order of Saint Jerome*) gathers his finding of a manuscript of the *Lazarillo* in the cell of Juan de Ortega (Sigüenza, 145):

> It is said that while being a student in Salamanca [i.e., Juan de Ortega], as a young man, he had such a fresh and gallant ingenuity, that he created that little book that moves around titled Lazarillo de Tormes, where he shows in that humble subject his mastery of the Castilian language and the decorum of the people introduced with such singular artifice and grace, that it deserves to be read by those of excellent taste. The reason for this was the discovery of the draft in his cell, handwritten by him.
>
> (Dicen que siendo [i.e., Juan de Ortega] estudiante en Salamanca, mancebo, como tenía un ingenio tan galán y fresco, hizo aquel librillo que anda por ahí, llamado Lazarillo de Tormes, mostrando en un sujeto tan humilde la propiedad de la lengua castellana y el decoro de las personas que introduce con tan singular artificio y donaire, que merece ser leído de los que tienen buen gusto. El indicio desto fue haberle hallado el borrador en la celda, de su propia mano escrito).

Although a draft was indeed found in the friar's cell, the circulation of handwritten copies was a common practice during the Spanish Golden Age (Botrel and Salaün). The claim that Father Ortega was the author is hard to sustain as the draft does not seem to be enough proof: it could have been the original as much as a handwritten copy or some annotated summary made by Juan de Ortega.

More than three centuries had to go by until the French hispanist Marcel Bataillon revisited the candidacy of Father Ortega, finding a satisfactory explanation for the anonymity of the *Lazarillo*. Friar Juan de Ortega received the habit in the Salamancan municipality of Alba de Tormes, and soon was chosen by King Charles V, Holy Roman Emperor, as archbishop of Chiapas in Mexico. He later became General of the Hieronymites from 1522 to 1555, which according to Bataillon, would sufficiently and objectively explain the reason of his not signing the little book around its publication in 1554.[8] Supporters of Bataillon's candidate include Claudio Guillén or Antonio Alatorre, who in 2002, and as a very final assertion, stated that «there is nothing comparable to the testimony of Friar José de Sigüenza» («No hay nada comparable al testimonio de fray José de Sigüenza»), suggesting that his sole mention was enough evidence (Alatorre, 447). It is likely that his statement was based on the idea defended by Bataillon that a book of the tone and kind of the *Lazarillo* would not be gratuitously attributed to a Hieronymite Friar. However, as noted by Francisco Rico, it is hard to know whether Father Sigüenza was even referring to the *right* Juan de Ortega (Anónimo ed. Rico, 120).

A couple of years after the proposal of Friar Juan de Ortega, another name took the centre stage and has probably been the most studied candidate ever since. In 2010 Alexander Sandy Wilkinson found editions of the *Lazarillo* made in 1599 in Zaragoza (Spain) by Juan Pérez de Valdivieso, and in 1600 in Rome (Italy) by Antonio Facchetti; both attributed the book to the diplomat and Governor of Grenade Diego Hurtado de Mendoza.[9] Surprisingly, these references went unnoticed, as it was only after his mention in the

---

8.– To this respect see the works by Marcel Bataillon (*El sentido*; *Novedad y fecundidad*).

9.– Following the citation in Corencia Cruz (16); see Wilkinson (652 and 414).

*Catalogus clarorum Hispaniae scriptorium* that the candidacy of the poet became popular. The Flemish bibliographer Valerio Andrés Taxandro wrote his *Catalogus* in 1607,[10] and in it he said that Diego Hurtado de Mendoza «owned a rich library of Greek authors, that he gifted to King Philip II of Spain on his death. He [*i.e.*, Diego Hurtado de Mendoza] also wrote romance poetry and the book of entertainment titled *Lazarillo de Tormes*» («Poseía rica biblioteca de autores griegos, que dejó al morir a Felipe II. Compuso también poesías en romance y el libro de entretenimiento llamado *Lazarillo de Tormes*») (Anónimo ed. Cejador y Frauca). A year later the Jesuit Andrés Schott also supported the attribution in his *Hispaniae bibliotheca*: «It is thought that the *Lazarillo de Tormes* is a work of his, book of satire and entertainment of his time as a student of civil law in Salamanca» («Se piensa ser obra suya el *Lazarillo de Tormes*, libro de sátira y entretenimiento de cuando andaba estudiando derecho civil en Salamanca»).[11] Accepting the attribution as true, Tomás Tamayo de Vargas confirmed it again in his *Collection of books the biggest that Spain has ever seen in its language up to 1624* (*Junta de libros la mayor que ha visto España en su lengua hasta 1624*): «Book of the most ingenious of Spain, and I do not know if in the foreign nations there is another of similar finesse in its subject. Valladolid by Luis Sánchez. 1603. 16th. Usually it is attributed this gracious birth to the ingenuity of Mr. Diego de Mendoza» («Libro de los mas ingeniosos de España, i no sè si en las naciones estranjeras hai otro de igual festividad en su assumpto. Valladolid por Luis Sanchez. 1603. 16°. Communmente se atribuie este graciosissimo parto al ingenio de D. Diego de Mendoza»).[12] Nicolás Antonio also contributed to the diffusion of Hurtado de Mendoza as the author, although he never completely rejected the previous candidate, Friar Juan de Ortega.[13] Despite the vague explanations, based mostly on the lack of evidence against him and some biographical similarities between him and Lázaro's life, the attribution proved to be extremely popular. For about three centuries book catalogues all over Europe recorded Diego Hurtado de Mendoza as the author of the *Lazarillo*.

The first serious criticism against this authorship came from another French hispanist, Alfred Morel-Fatio, who in 1888 proposed a new candidate, Juan de Valdés —to whom we will come back later—, giving a start to the modern attribution wars of the *Lazarillo* (Morel-Fatio, 112-76). Alfred Morel-Fatio's main claim was alluding to the number of attributions granted to Hurtado de Mendoza that were based solely on his reputation as *enfant terrible*, literarily speaking. All the objections against Hurtado de Mendoza that Morel-Fatio formulated were refuted several times by Ángel González Palencia.[14] The Arabist and literary critic also noted some analogies between the uninhibited character of the *Lazarillo* and the tone employed by Hurtado de Mendoza in his private correspondence; albeit of

---

10.– Some authors argue that Valerio Andrés Taxandro was a pseudonym of Andrés Schott, see for example Francisco Calero («Vives y el *Lazarillo*»).

11.– The citation can be found virtually in any edition of the *Lazarillo* or study about its authorship, we use Rico's 2011 edition. However, the original, in latin, belongs to Andreas Schott.

12.– As edited in her PhD thesis María Cristina González Hernández (401).

13.– Although the edition preserved is from 1783, Nicolás Antonio wrote it in 1672.

14.– See his edition of the *Lazarillo* (Anónimo ed. González Palencia; «Leyendo el Lazarillo» 3-39). From 1941 to 1943, and together with Eugenio Mele, they also collected, edited, and published the works and biography of Diego Hurtado de Mendoza (González Palencia and Mele).

acknowledging the stylistic dissimilarities to later conclude that the attribution «is not unlikely» («no es improbable») (González Palencia, «Leyendo el Lazarillo» 36):

> It shall not be emphasized the stylistic aspect of the Lazarillo with purposes of comparison to the works by Mendoza: the dry, short, and succinct style of the Lazarillo agrees to that of Mendoza's letters and others prose works of him. However, this aspect should not be highlighted, considering that such writings, as a post data, and for commenting news or events, had to be written inevitably hastily, in a shortened, fast, and edgy way.
>
> (No puede hacerse gran hincapié en el aspecto estilístico del Lazarillo para compararlo con los escritos de Mendoza: el estilo seco, cortado y conciso del Lazarillo concuerda con el de estas cartas de Mendoza y con otras obras en prosa suyas. Pero acaso no se le pueda y deba dar gran valor a este punto, teniendo en cuenta que tales escritos, en forma de postdata, y para comentar una noticia o un suceso, habían de escribirse forzosamente de prisa, en forma abreviada, rápida y nerviosa).

The ideas presented by the critic laid the foundations for other scholars, specially for Erika Spivakovsky. Unlike González Palencia, who believed that Hurtado de Mendoza wrote the *Lazarillo* when still young —following on Andrés Schott's footsteps—, the American researcher gave a much later date for the conception of the book, effectively defending that the little novel was written in 1553, which coincided with the mature years of Diego de Mendoza. «We have few notices about Mendoza during 1553-1554 [writes Erika Spivakovsky]. Yet so much is known that, remarkably, he did not only had just the time and opportunity to do some writing for his own pleasure, but it seems to have been, in fact, the only period of his active middle years when he might have found a few weeks of complete leisure to perfect such as masterpiece» («The Lazarillo» 273). The sentence summarizes her most important contribution to the debate: a noticeably precise series of parallels drawn between Hurtado de Mendoza's life and the fortunes and adversities of Lázaro de Tormes and those whom he found in his path. The analogies are numerous, *e.g.* between Pope Paul III and the Blindman, the Sienese conspirator Amerigo Amerighi and the Cleric, or Charles V and the young Squire.[15]

As convincing as it may sound, without factual evidence the intellectual exercise by Spivakovsky, and the whole Diego Hurtado de Mendoza candidacy, falls exclusively on the realms of metaphor and hermeneutics. At least until 2010, when Mercedes Agulló claimed to have found the missing piece of the puzzle. The Madrilenian historian published a monograph detailing the testament and inventory of goods of Diego Hurtado de Mendoza, as recorded at his death by the administrator of his estate, Juan López de Velasco. In one of the drawers containing books of López de Velasco,[16] among other panniers belonging to Diego Hurtado de Mendoza, there was one that read: «a bundle of corrections made for the printing of *Lazarillo* and *Propaladia*» («Vn legajo de correçiones hechas para la ynpressión de *Laçarillo* y *Propaladia*») (Agulló y Cobo, *A vueltas* 44). The sentence, together with other surrounding historical circumstances, was sufficient for

---

15.– See Spivakovsky («¿Valdés o Mendoza?» 15-23), and her book *Son of the Alhambra*. Others such as Olivia Crouch and Charles Vincent Aubrun also supported the idea, but added little to the discussion (Crouch, 11-23; Aubrun, 240).

16.– The drawer was part of López de Velasco's will, but Agulló defends that since everything that was in the drawer belonged to Hurtado, and Velasco was the executor of Hurtado's will, the drawer belonged to Hurtado as well.

Mercedes Agulló to cautiously relaunch the old candidacy of the diplomat.[17] The finding must not be minimized though, as it is the best documentary evidence to date. However, it is also true that all the documents were released as part of Juan de Valdés' will, the lawyer who made the inventory of Juan López de Velasco's fortune, which in turn included that of Hurtado de Mendoza. Although Agulló argues that Diego Hurtado de Mendoza's files were bundled together and distinguishable from those of the executor of his will, the fact that López de Velasco was the person in charge of the *Lazarillo Castigado* makes the statement gain some uncertainty: the corrections as such are lost and another book is mentioned along with the *Lazarillo*. Strong reactions and criticism came shortly after Agulló published her work. In the same year several essays appeared refuting her findings, all of them mostly centered around the aforementioned questions about the impossibility of stating much about Diego Hurtado de Mendoza's authorship: it is not clear why Hurtado de Mendoza would have made corrections to Bartolomé Torres Naharro's *Propalladia* (*Propaladia*); and it might make more sense that the corrections were made by the censor Juan López de Velasco himself prior to the preparation of his expurgated edition.[18] And although some openly supported Mercedes Agulló,[19] she defended herself in a second article published a year later. The historian suggested then that López de Velasco, in order to work on his expurgated edition, called upon Hurtado de Mendoza to provide him with the right corrections, thus being the nature of the *legajo* (bundle) referred in López de Velasco's documents. Agulló uses the attribution to explain the nature of a book: Hurtado de Mendoza sent a letter to his nephew, to which said book was attached. In this letter, Hurtado asked his relative to hand in the book to Philip II, then still a young prince, and to warn the future king not to take the book too seriously, as Hurtado did not want to be on the spotlight on account of the told «necedades» («follies»).[20] She leaves, however, other mysteries to the reader, such as the reason for the absence in Hurtado's library of many of the books that are believed to have influenced the *Lazarillo*, arguments sometimes used against Hurtado de Mendoza's candidacy but that require a more thorough research.[21]

In the long process of debating against Diego Hurtado de Mendoza's authorship, other names were brought to light. In 1867 José María Asensio published previously unseen work by the dramatist, jurist, and Toledo born, Sebastián de Horozco. *Representación de la historia evangélica del capítulo nono de San Juan* (*Representation of the evangelical history of* *the ninth chapter of Saint John*) exhibits —according to José María Asensio— some similarities between a blind man guide character named Lázaro and the protagonist of the *Lazarillo* (*Sebastián de Horozco* 46). Julio Cejador y Frauca, after rejecting other authors such as the Valdés brothers, Cristóbal de Villalón, or Lope de Rueda, took José María Asensio's suggestion and supported it with ever more similitudes, matches of themes and characters, and some biographic coincidences: «It was written, by whoever, in Toledo, even though [the author] sets the beginning of the action in Salamanca and appears himself knowledgeable about that city [...], this points out [...] entirely to Sebastián de Horozco» («Escribiolo, fuera quien fuera, en Toledo, aunque ponga el comienzo de la acción en Salamanca y se muestre bien enterado de aquella ciudad [...] esto compete [...] de lleno a Sebastián de Horozco»).[22] The inclination towards popular sayings in Horozco's works ended up convincing Cejador of the candidacy of the Toledan. However, just a year later Emilio Cotarelo started the publication of Sebastián de Horozco's *Refranes glosados* (*Glossed sayings*), where the candidacy of the jurist was solidly rejected and abandoned by everyone else ever since (Horozco ed. Cotarelo). It was forty years later when Francisco Márquez de Villanueva brought this candidacy back without adding much to the debate; his name and authority, however, would suffice for many others to also rethink about and support it. Up to two times Francisco Rico rejected Horozco's candidacy arguing that the use of the language was very different between the two books. While it seems to be evident that the Toledan took some inspiration from the *Lazarillo*, Rico states that the rich linguistic inventory and expressive power in the little book surpasses any effort made in the *Representación*, which accents the vulgarity of folk speech and exposes a lack of narrative imagination (Márquez Villanueva, «Sebastián de Horozco» 253-339; Anónimo ed. Rico 1987 and 2011).[23]

Shortly after José María Asensio proposed Horozco, Morel-Fatio, based on the anticlerical tone of the little book, pointed towards the circle of humanists surrounding the Valdés brothers (Morel-Fatio, *Recherches* 164-166). From there, some decades later Manuel J. Asensio built his case in favor of the younger brother,[24] the reformist Juan de Valdés, placing the writing of the *Lazarillo* near Escalona and Toledo around 1525 (Asensio, *La intención religiosa*; Asensio, «El Lazarillo» 101-28). As Asensio himself defended, his prudent proposal never pretended to be a conclusive argument to justify the attribution, but rather a clue for others to follow. Joseph V. Ricapito took the lead on this matter when in 1976 he supported «a very risky hypothesis» («una hipótesis arriesgadísima») of the attribution of the *Lazarillo* to the older of the Valdés brothers, Alfonso, chancellor and Royal Secretary of Indian Letters of Emperor Charles V. In Ricapito's own words, if Alfonso de Valdés was not the author, «it had to be someone suchlike him and someone who belonged to the same intellectual circles» («tuvo que ser alguien semejante a él y alguien que perteneciera a los mismos círculos intelectuales») (Anónimo ed. Ricapito). More recently, after carefully editing the *Diálogo de las cosas acaecidas en Roma* (*Dialogue of the things occurred in*

*Rome*) and the *Diálogo de Mercurio y Carón* (*Dialogue of Mercury and Charon*) —both apparently wrongly attributed to Juan de Valdés until the end of 19th-century and 1925, respectively—, Rosa Navarro Durán came into the discussion to also back up the candidacy of Alfonso de Valdés (Navarro Durán, *Alfonso de Valdés*). The Catalan philologist carried out a detailed study of the books that influenced the author of the *Lazarillo*, whoever that might be, and the readings that inspired Alfonso de Valdés in his works. Finding that both the *Diálogos*'s and our little book's author shared the same literary roots, Navarro Durán concluded that the writers must have been the same person. The problem with this strong assumption is that it implies a very early date for the conception of the *Lazarillo*, as the older of the Valdés brothers died of the plague in Vienna in 1532. Conveniently, all the books that apparently served as source for Alfonso de Valdés in the writing of the *Lazarillo* were available before that date (works such as *La Celestina* [*Tragicomedy of Calisto and Melibea*] by Fernando de Rojas, the *Propalladia* by Torres Naharro, the anonymous *Comedia Thebaida* [*Comedy called Thebais*], *La lozana andaluza* [*The lusty Andalusian woman*] by Francisco Delicado, or even the *Relox de príncipes* [*Watch of Princes*] by Antonio de Guevara). And when not, as Francisco Rico noted in relation to the *Dichos graciosos de españoles* (*Funny sayings of Spaniards*) collected by Chevalier or the *Baldus* by Folengo, Navarro Durán interprets it as the *Lazarillo* influencing other works, instead of being influenced by them (Navarro Durán, *Lazarillo*; «Lazarillo de Tormes»).

As noted many times, the last paragraph of the prologue in the *Lazarillo* does not seem to correspond with the authorial voice present in the rest of the little book.[25] Navarro defends that two different discourses can be identified: one coming from the author himself, and the other from Lázaro, the character, narrating «the case» («el caso») to «Your Grace» («Vuestra Merced»). The philologist also points in the direction of a supposedly disappeared folio that used to accompany all literary works in the 16th-century, and that would split the prologue and the body, separating in practice the two distinct narratives. She imagines an *Argumento* (Argument) of erasmist nature articulated upon the secret of confession, in her opinion key for the correct understanding of the little book and she precisely defends that it was because of this that the page was torn off. Furthermore, and exhibiting a laudable creative dexterity, she proposes that «Vuestra Merced» is in fact a woman, who having confessed to the Archpriest of San Salvador, gets worried after discovering his *amancebamiento* (de facto relationship) with a maidservant married to no one less than a town crier of the wines of Toledo, our own Lázaro de Tormes, to whom «Vuestra Merced» asks for explaining the case and dispel her doubts (Navarro Durán, «El caso» 3-9; *La verdad*). While this adds little to the question of the author, her reflections resonated with some scholars who encumbered her at the peak of erudition.[26] Others carried out studies dismantling every aspect of her theory. Despite the efforts of the Catalan framing the *Lazarillo* as erasmist to harmonize with the style of Alfonso de Valdés, and the recent support in 2010 by the pioneer of the attribution (Ricapito, «Further Comments» 95) —possibly aimed by the popularity reaped by Navarro Durán—, there are still strong reasons against Valdés. To cite a few: the lack of solid linguistic con-

---

25.– Others argue that the prologue must be read in the last place, as a final treatise (Lázaro Carreter 134; Sieber).

26.– In favor we can mention Juan Goytisolo (sec. 26).

cordances, the difference in style and genre (Alfonso de Valdés wrote mostly theological works), the aforementioned gap between the writing date and the first known editions of 1554, and the fact that the second part of the little book (which Navarro grants to Hurtado de Mendoza) starts with Lázaro enrolled to the war in Argel in 1541.[27]

At the beginning of the 20th-century, Fonger de Haan related the existence in 1538 of a town crier of Toledo named Lope de Rueda. Julio Cejador y Frauca accounts for the fact and, as part of his arguments in favour of Sebastián de Horozco, rejects what he considered to be a weak proposal for the authorship of the *Lazarillo* (Anónimo ed. Cejador y Frauca). The discovery led Fred Abrams to believe that the town crier was in fact the Sevillian actor and author of *entremeses*, Lope de Rueda. After analyzing the thematic and stylistic similarities as well as the concordances between the little book and the plays by Lope de Rueda, the American suggested that the actor could be the wanted author (Abrams, 67). However, a later study on town criers by Jaime Sánchez Romeralo revealed that the Lope de Rueda from Toledo and the author of plays were different persons, which was considered by Rico as the final piece of evidence to stop supporting the candidacy of the playwright. Years later Alfredo Baras Escolá still defended the similarities between the works of the Sevillian Lope de Rueda and the *Lazarillo*, based on the «eleven motifs or situations usually employed by the dramatist Lope de Rueda and that happen with precision in the novel [*i.e.*, the *Lazarillo*] in the form of sequences» («once motivos o situaciones a que suele recurrir Lope de Rueda dramaturgo y que se cumplen con exactitud en la novela incluso en forma de secuencias») (Sánchez Romeralo, «*De Lope de Rueda*» 671-675; Baras Escolá, «Lazarillo y su autor» 6), but with the scholar having failed to express them clearly, no one seems to have supported the actor's authorship ever since. Rico vehemently rejects the hypothesis: «the same alleged reasons that later on would be used in defense of this idea lead to discard them without hesitation» («las mismas pretendidas razones que posteriormente se han querido alegar en defensa de tal idea inducen a descartarla sin vacilaciones») (Anónimo ed. Rico, 40).

During the second part of the 20th-century other names were proposed although none of them enjoyed enough support afterwards. In 1955, based on the idea of the author being a recognized intellectual and humanist in Spain at the time, Arturo Marasso raised the possibility of the professor and latinist Pedro de Rhúa (Marasso, 74). His argument was based on an alleged aversion between Pedro de Rhúa and Friar Antonio de Guevara. In that sense, the *Lazarillo* would merely be a parody of Guevara's style, in particular of his *Epístolas familiares* (*Family epistoles*). The Argentinian also highlighted the erasmist and knowledgeable tone used by de Rhúa in his letters. The criticism against Antonio de Guevara is somewhat similar to the general indictment towards the clergy from Soria that can be found in the *Diálogos de Mercurio y Carón* (Corencia Cruz). To this respect, Fernando Calero contributed supporting the candidacy of de Rhúa as the author with a rather particular approach (Calero Calero, «Homenaje» 26):

---

27.– Notable critics against the thesis of Navarro Durán include Alatorre (*Los denigradores*; «El Lazarillo» 143-51), Féliz Carrasco («Lazarillo» 9; «¿Errata o *lectio difficilior*?» 23), Francisco Márquez Villanueva («*El Lazarillo y sus autores*» 137), Valentín Pérez Vénzalá («El Lazarillo» 46), Marco Antonio Ramírez López («Fortunas» 43), Pedro Martín Baños («Nuevos asedios» 2).

What a sharp nose Marasso had! Because the hidden author of the Lazarillo was indeed the Bachiller Rhúa. [...] It seems highly significant that in all Spanish literature [the expression «lana caprina» (goat wool)] was only used in the Letters of Rhúa, and from there the concordance with Vives [in regards to his *De concordia et discordia in humano genere*] gained an incontrovertible evidential value. If we join this concordance to the other previous two, there is no shadow of a doubt that Rhúa and Vives are the same person.

(¡Qué fino olfato literario tuvo Marasso! Porque, efectivamente, el oculto autor del Lazarillo fue el Bachiller Rhúa. […] Resulta altamente significativo que en toda la literatura española sólo sea utilizada [la expresión 'lana caprina'] en las Cartas de Rhúa, y de ahí que la concordancia con Vives [en su *De concordia et discordia in humano genere*] adquiera un valor probatorio incontrovertible. Si unimos esta concordancia a las dos anteriores, no puede caber la más mínima duda de que Rhúa y Vives son la misma persona) (qtd. in Sánchez Ferrer, *Los padres*).

Aldo Ruffinato also found Brenes' hypothesis to be evocative and compelling.[28] Unfortunately, the profile of the author drawn by Marasso lacks bibliographical support and factual certainties to rely on.

In his 1964 essay on the interpretation and attribution of *Lazarillo*, Aristide Rumeau proposed the latinist Hernán Núñez de Toledo as the author (Rumeau). His comparison between the little book and *Las trescientas del famosísimo poeta Juan de Mena con glosa* (*The three hundred of the universally known poet Juan de Mena with glosa*) by the disciple of Nebrija, relied on linguistic and tone similarities, although these were not compelling enough to raise the support of other scholars. Not a stronger candidate is Fernando de Rojas, proposed by Howard Mancing in 1976. The American researcher based his hypothesis on the ability of the alleged author of *La Celestina* to criticize the social establishment, and on his nature of *converso* (convert), which supposedly granted him an agnostic or anti-clergy background to write the *Lazarillo* (Mancing, 47-61). The Royal Secretary Gonzalo Pérez was also proposed by Dalai Brenes Carrillo in a series of studies started in 1986. Brenes interprets that the translator of *La Vlixea de Homero* (*The Odyssey of Homer*) wrote the little book as a sort of *roman à clef* about the life in the court of Charles V, where Laźaro is a «combined anti-thesis of the young Telemachus and the astute Ulysses of the gimmicks» («combinada antítesis del joven Telémaco y el astuto Ulises de las tretas.») (Brenes Carrillo, «Lazarillo» 43; «Vlixea» 104). In the process, Brenes identifies the addressee of «V.M.» as «Vuestra Majestad» (Your Majesty), and establishes other parallels between characters of the little book and real ones surrounding the milieu of the Emperor (Hurtado de Mendoza, Fernando de los Cobos, Gattinara, Enciso, Sílice, etc.) (Brenes Carrillo, «¿Quién es V.M.?» 73-88).[29] Other minor attributions, at least in terms of supporters and evidence, include the dramatist Bartolomé Torres Naharro, author of *La Propalladia*, who, according to Alberto M. Forcadas, shares certain similarities with the *Lazarillo* (Forcadas, 48). Furthermore, suggested for the first time by Cejador (Anónimo ed. Cejador y Fracuca), Juan Maldonado was more recently supported by Clark Co-

28.– In his *Introduction* of his edition of the *Lazarillo*.

29.– Curiously, in his *Un par de vueltas más*, 2011, Agulló claimed that «V.M.» was in fact referring to Gonzalo Pérez himself.

lahan and Alfred Rodríguez in 1995. Arguing that although the humanist and friend of Erasmo only wrote in Latin, the little book presented several thematic and stylistic correspondences, supported by the common style used by Maldonado, *i.e.*, the autobiographical monologue.[30]

Almost all previous candidates were rejected in 2003 by Francisco Calero, who staunchly defended Juan Luis Vives, the illustrious Valencian pedagogue and philosopher, as the author of the little book. Despite his thorough analysis of up to 151 (sic) thematic, stylistic, and linguistic concordances, more than enough to *incontrovertibly* settle the problem once and for all —in Calero's words—, the candidacy still does not feel sufficiently strong. Drawing on the work of other *lazarillistas*, the philologist seems to arbitrarily use the arguments that could benefit his thesis while rejecting those that do not, *e.g.*, «[Américo Castro] previously defended the Jew origin of Luis Vives. It is true that he did not propose him as the author of the *Lazarillo*, but it is also true that he was in the right direction» («Al igual que en los casos citados, también acertó en este A. Castro, quien con anterioridad había defendido el origen judío de Luis Vives. Es cierto que no llegó a postularlo como autor del Lazarillo, pero también lo es que estaba en la dirección correcta») (Calero, «Luis Vives»; *Juan Luis Vives, autor* 46). Besides the concordances, Calero's arguments rely on the conviction that the erasmist also wrote in the Castilian language, although Vives was known and laureated for his Latin works in several and complex matters such as hunger, poverty, charity, mercy, spirituality, or morality. In order to further support his claims, Calero builds on Ricapito's arguments to sustain Alfonso de Valdés' authorship and twists them to favour his candidate (Calero, «Homenaje» 65). Likewise, he supports Navarro Durán's thesis in one important aspect: the author of the *Diálogo de las cosas acaecidas en Roma* and *Diálogo de Mercurio y Carón* must be the author of the Lazarillo. Coincidentally, Calero has published several works that allegedly demonstrate that the *Diálogos*, together with other important works of the time, were all written by Juan Luis Vives. In his zeal, the philologist passes over the inquisitorial documental proof referred to by Bataillon that attributes both the *Diálogos* to Alfonso de Valdés. And while we acknowledge the similarities between the *Diálogos* and the *Lazarillo*, the topics and expressions alluded by Calero to defend his thesis as unequivocal were actually platitudes among the intellectual circles at the time. The early dead of the forefather of modern psychology in 1540 does not help in either case. More recently the attribution to Vives was supported by Marco Antonio Coronel Ramos in 2012, without really adding much (81), and criticized again in 2014 in a review of Calero's theory by Encarna Podadera, editor of a critical edition of the second part of the little book (13-24).

The 21st Century also brought the first authorship attributions complemented and supported by computational means. In order to delimit the profile of the author, in 2003 José Luis Madrigal drew his attention to the circle of intellectuals surrounding Alejo de Venegas.[31] The grammarian from Toledo wrote *Las diferencias de libros que ay en el Universo* (*The different books existing in the Universe*) in which the «libro racional» («rational book») covers the topic of poverty with influences from Erasmo's *Moria* and a general

---

30.– See Colahan and Rodríguez (289-311), and to a lesser extent Warren Smith, Clark Colahan, and Alfred Rodríguez (160-234).

31.– Vaguely proposed as well by Ruffinatto («Lázaro González Pérez» 3).

tone inspired by Apuleyo's *Asno de oro* (*The Golden Ass*). The evident erasmist point of view inspired Madrigal to conclude that the author of the little book had to be a disciple of Venegas, and if not from Toledo at least a great connoisseur of the place. After discarding other authors of the same environment, Madrigal found in Francisco Cervantes de Salazar the candidate that fitted the profile («Estudio de atribución» 9-13; «Cervantes de Salazar» 3). Translator of Juan Luis Vives, Fernán Pérez de Oliva, and Luis Mejía among others, Cervantes de Salazar moved to Mexico possibly inspired by the opportunity to found the Pontifical University of Mexico. There he started to sign his works with the Latin version of his name, *Franciscus Cervantes Salazarus*, in which Madrigal believed to find an anagram with the name Lázaro (*saLAZARUS ∼ LAZARO*) that would prove the authorship, albeit recognizing himself that «attributions based in possible anagrams usually have the same credibility that the prophecies of Nostradamus» («Las atribuciones basadas en posibles anagramas suelen tener normalmente la misma credibilidad que las profecías de Nostradamus») (Madrigal, *Autor del Lazarillo*). To further support his candidate and his circumstantial evidence, Madrigal tried to identify what he called the *modus scribendi* of the author, a sort of fingerprint that comprises the set of features that supposedly defines the style of an author univocally. From the electronic versions of texts available in repositories such as CORDE,[32] and using as discriminator the coincidences between the *Lazarillo* and Cervantes de Salazar's *Crónica de Nueva España* (*Chronicle of New Spain*), Madrigal built a method upon four opinionated levels of similarity (groups of words, idioms, peculiar syntactic turns, and other complex syntactic constructions). After applying his technique to other contemporary works to see which ones kept the highest number of similarities, Cervantes de Salazar's works were stylistically closer to the *Lazarillo* that any other work. During the process Madrigal acknowledged he had not used more modern and current approaches to authorship attribution, which weakens the credibility of his proposal although not of his methodology. In fact, five years later, with more evidence and slightly improved methods, Madrigal was forced to abandon the candidacy of the Toledan and welcome the jurist Juan Arce de Otálora, author of the *Coloquios de Palatino y Pinciano* (*Colloquia of Palatino and Pinciano*) (Madrigal, «Notas» 137-236). The palinode, strongly criticized by scholars such as Francisco Calero («Los Coloquios» 65), downplayed the issue arguing that during the research, the corpus he had access to was limited, and that he realized that the author did not necessarily need to be a member of the chosen corpus —a problem usually referred to as the *open-set problem* and that characterizes the attribution of the *Lazarillo*. Nevertheless, Madrigal continued to employ keywords in context (KWIC) concordances to further support Arce de Otálora's candidacy, insisting as well on another anagram he found («LAZARO DE TOR(M)(E)S ARZE DE OTALOR»), and the suggestive coincidence between Lázaro's surnames (González Pérez) and Arce de Otálora's grandparents surnames (Fernand González and Juan Pérez).[33] In the same year of 2010, Alfredo Rodríguez López-Vázquez supported and rejected the candidacy of the author of the *Coloquios*, to later propose Friar Juan de Pineda («El Tractado» 259-72; «Una refutación» 313-34). His theory was based on the

---

32.– Banco de datos (CORDE), 2007, October 30 2015 ‹http://www.rae.es›.

33.– In the Hispanic tradition is habitual that people have two surnames, the first coming from the first surname of the father, and the second from the first of the mother (Madrigal, «De nombres y lugares» 89-118).

same statistical methods and biographical similarities used by Madrigal, and followed the lead left by José Luis Ocasar, who edited the *Coloquios* some years before and in a later study did not confirm nor deny Arce de Otálora as the author (Ocasar, 873-888).

Inspired by Madrigal and Rodríguez López-Vázquez, and by means of his genetic-literary approach,[34] Ocasar mentioned a possible collaboration in the little book by Friar Juan de Pineda. The editor of Otálora's *Coloquios* highlighted that way the alleged important role of the multiple authorship around the mid 16th-century, previous to the strict rules imposed by the Church in terms of the signing of books, and raised the possibility of *Lazarillo* being the product of the collaboration between several authors. Although the analysis of multiple authorship may be increasing in importance and interest in recent years it was once considered a sort of joke, as gathered by Francisco Rico in relation to Francisco de Avellaneda's 1675 famous *Loa por papeles* (*Loa for the papers*):[35]

> I do not ignore that Thou knows,
> as [Thou] nothing ignores,
> that the Lazarillo de Tormes
> six lads, just like that,
> wrote in two days,
> as that is the utter count.
> (No ignoro que Vos sabéis,
> puesto que nada ignoráis,
> que al Lazarillo de Tormes
> seis mozos, sin más ni más,
> escribieron en dos días,
> que esta es la cuenta cabal).

At this side of the spectrum at which the author is belittled in favour of the many interpretations and meanings that the anonymity has to offer, some scholars such as Robert Fiore consider the authorship of the little book vital for its understanding: «the author, who undoubtedly wished to remain anonymous, has had his wishes fulfilled. Not only does the author remain unknown today, but his narrator is obscured, and his point of view is so shrouded by irony that it is not obvious to readers and critics» (Anónimo ed. Fiore, 714). In the same line, Américo Castro suggests that the anonymity of the *Lazarillo* is an essential part of the text itself:[36]

> We should realize, however, that this anonymity is not an accident, nor an omission, but an essential aspect of the literary reality of the book. If we take the fact of this anonymity as a point of departure, we may penetrate the book more deeply and enjoy it better than through mere appeasement of our curiosity about the author's name.

In his latest edition of the *Lazarillo* to date, and after thoroughly discrediting all other candidates, Francisco Rico takes for certain that the author was indeed a man named

---

34.– Roughly, a genetic-literary analysis is the study of the differences and similarities between the editions of a text.

35.– Rico's *Lazarillo* (115-128), where he also gathers the attribution made by Dr. Locker, Dean of Peterborough, to a group of Spanish bishops traveling to the Council of Trent.

36.– See Américo Castro's introduction in Williams Harry Franklin and Hesse Everett Wesley.

«Lázaro de Tormes». In order to argument in favor of the apocryphal character of the book, Rico maintains that the game-changer aspect of the *Lazarillo* was a new kind of fiction, one that the audience was not yet ready to experience: «readers faced the book as pure 'truth' and ended up finding a 'lie' that established an admirably new genre of 'fiction'» («los lectores acometían el libro como pura 'verdad' y acababan encontrando una 'mentira' que instauraba un género de 'ficción' admirablemente nuevo») (Anónimo ed. Rico, 115-128). According to Rico, not all readers were capable or in a position to decypher the fictionality introduced in the *Lazarillo*. This same complexity, together with the structural necessity of the author for anonymity, also led Fernando Rodriguez Mansilla to think about the author as an undercover moralist, not as a professional writer, who only wrote one little book in his entire life (Rodriguez Mansilla 235). We have a precedence in Fernando de Rojas' *La Celestina*. If this were true, as Rico points out, any internal analysis of the little book would have been futile. Therefore, we will work from the assumption that its true author, as slippery and elusive as he may seem, wrote more than only one book, even if that were a masterpiece such as the *Lazarillo*.

According to the aforementioned list of the most frequently proposed authors, we have created a table that summarizes the candidates in terms of support by scholars and sorted by year of contribution (see table 1), as well as a chronology of the candidates, when they were proposed, by whom, who criticized them, and when they were criticized (see also table S1 in the supplementary materials, henceforth: SM).[37]

Table 1: List of plausible candidates as mentioned in this study, by year of proposal. For each author a chronological list of scholars supporting and rejecting the hypothesis is shown.

| Author | Supported by | Year | Criticized by |
|---|---|---|---|
| Juan de Ortega | José de Sigüenza | 1605 | |
| | | 1624 | Tomás Tamayo de Vargas |
| | Marcel Bataillon | 1954 | |
| | Claudio Guillén | 1966 | |
| | " | 1988 | |
| | Antonio Alatorre | 2002 | |

37.– Good summaries can be found in Rico's 2011 edition, and Joaquín Corencia Cruz. Rico's 2011 edition is not included in this table as he basically discredited all the authors ever proposed. He stays neutral while the same edition reads «Lázaro de Tormes» as the author.

| | | | |
|---|---|---|---|
| Diego Hurtado de Mendoza | Valerio Andrés Taxandro | 1607 | |
| | Andrés Schott | 1608 | |
| | Tomás Tamayo de Vargas | 1624 | |
| | Nicolás Antonio | 1873 | |
| | | 1888 | Alfred Morel-Fatio |
| | Ángel González Palencia | 1943 | |
| | Eugenio Mele | " | |
| | Erika Spivakovsky | 1961 | |
| | Olivia Crouch | 1963 | |
| | Charles Vincent Aubrun | 1969 | |
| | Erika Spivakovsky | 1970 | |
| | Mercedes Agulló | 2010 | Javier Blasco |
| | Jauralde Pou | " | Rosa Navarro Durán |
| | | " | José Luis Madrigal |
| | | " | Rodríguez Mansilla |
| | Mercedes Agulló | 2011 | |
| | Reyes Coll-Tellechea | " | |
| | Joaquín Corencia Cruz | 2014 | |
| Sebastián de Horozco | José María Asensio | 1867 | |
| | Julio Cejador y Frauca | 1914 | |
| | | 1915 | Emilio Cotarelo |
| | F. Márquez Villanueva | 1957 | |
| | José Gómez-Menor Fuentes | 1973 | |
| | Jaime Sánchez Romeralo | 1978 | |
| | Fernando González Ollé | 1980 | |
| | | 1987 | Francisco Rico |
| Juan de Valdés | Alfred Morel-Fatio | 1888 | |
| | | 1914 | Julio Cejador y Frauca |
| | Manuel J. Asensio | 1959 | Erika Spivakovsky |
| | " | 1960 | |
| | " | 1992 | |
| Lope de Rueda | Fonger de Haan | 1901 | |
| | | 1914 | Julio Cejador y Frauca |
| | Fred Abrams | 1964 | |
| | | 1980 | Jaime Sánchez Romeralo |
| | | 1987 | Francisco Rico |
| | Alfredo Baras Escolá | 2003 | |
| | | 2006 | Francisco Calero |
| Pedro de Rhúa | Arturo Marasso | 1955 | |
| | Francisco Calero[37 *] | 2008 | |
| Hernán Núñez Toledo | Aristides Rumeu | 1964 | |
| | | 1987 | Francisco Rico |

37 *.– Francisco Calero does not properly support the authorship of Pedro de Rhúa per se, but him being the same person than Juan Luis Vives («Homenaje» 26).

| Alfonso de Valdés | Joseph V. Ricapito | 1976 | |
| | Rosa Navarro Durán | 2002 | Antonio Alatorre |
| | " | 2003 | « |
| | Juan Goytisolo | " | Félix Carrasco |
| | | 2004 | « |
| | | " | F. Márquez Villanueva |
| | | " | Valentín Pérez Venzalá |
| | Rosa Navarro Durán | 2006 | M. Antonio Ramírez López |
| | | 2007 | Pablo Martín Baños |
| | Rosa Navarro Durán | 2010 | |
| | Joseph V. Ricapito | " | |
| Francisco Cervantes de Salazar | José Luis Madrigal | 2003 | |
| | | 2008 | José Luis Madrigal |
| Juan Luis Vives | Francisco Calero | 2006 | |
| | M. Antonio Coronel Ramos | 2012 | |
| | | 2014 | Encarna Podadera |
| Juan Arce de Otálora | José Luis Madrigal | 2008 | |
| | Rodríguez López-Vázquez | 2010 | Rodríguez López-Vázquez |
| | | 2011 | Francisco Calero |
| | José Luis Madrigal | 2014 | |
| Juan de Pineda | Rodríguez López-Vázquez | 2010 | |

## Beyond Concordances

The analysis of texts sits at the core of the humanities. Identifying writing styles and the authors of anonymous or wrongly attributed texts has been of interest to scholars at least since the invention of the printing press, when the availability of and access to texts fostered comparative studies.[38] Text attribution studies fall into two different categories attending to the nature of the evidence used. Internal analysis deals with the ways in which language is employed, from syntactic occurrences to the use of expressions that are characteristic of a specific author, or stemmatics based on Lachmannian textual criticism. The purpose of the internal analysis is to find the fingerprints of the author, and in the process it assumes that each author maintains a *modus scribendi* —as Madrigal called it— that is characteristic of each author. On the other hand, external analysis focuses on the circumstances of the author and how they are reflected in the text. It tries to create a profile of the anonymous writer by looking into readings that influenced the text, the kind of content expressed, and even by tracing parallels between events told in the text and the real life events of the author. Historiography, hermeneutics and rhetorics are big parts of the external analysis. Most non-traditional authorship attribution studies, in opposition to those run by the domain experts, rely on the internal analysis of the texts in hand, and therefore assume some existence of a quantifiable individual's writing style.

Although authorship studies and their quantitative approaches predate computing, the introduction of computers made it easier and more affordable to analyze internal characteristics of texts and whole corpora (Love; Lord, 282; Mendenhall, 97-105). The

38.– See an introduction to the topic by Harold Love.

successful attribution made by Frederick Mosteller and David Wallace of the essays in *The Federalist* marked the start of modern authorship techniques powered by computers. Their method was based on the statistical analysis of a set of predefined characteristics —usually a list of words— known as style markers: features outside the conscious control of the writer that were supposed to quantify the writing style. Over time other features were added, such as sentence length, vocabulary richness, *magic* indices (such as the widely used Yule's Characteristic or Simpson's Index), hapaxes, character frequencies, and all sort of ratios.[39] However, the case of the *The Federalist* has been considered not to be a good representative of the larger problem of non-traditional authorship studies: Mosteller and Wallace had a very well delimited problem with a clear set of possible candidates and certainty about one of them being indeed the author. Criticism started to flourish after a period of popularity during which the results of authorship attribution techniques were even accepted at courts as experts' evidence.[40] Richard Bailey was the first to identify the necessary circumstances for authorship attribution in a forensic setting (1-20). More recently Efstathios Stamatatos considered (and extended) those as limitations of the techniques when faced with real life authorship attribution cases: long textual data of possibly very dissimilar styles, small candidate sets with 2 or 3 members, corpora not controlled by topic, and lack of objective evaluation criteria or benchmark data to assess the goodness of the methods («A survey» 56). These flaws would be mostly overcome from the 1990s onwards, when electronic texts became pervasive and machines started to be powerful enough to process large volumes of data. In turn, these advances made possible the development and maturation of disciplines such as information retrieval, machine learning, and natural language processing (NLP), from where authorship studies have borrowed and applied some of the methods (Stamatatos, «A survey» 56).

Other aspects also affect the credibility and accuracy of computer-based methods. There is evidence of language affecting the reliability of these techniques, mostly focused in English texts since its beginning, although some language-independent methods of attribution have later appeared as part of computational linguistics (Peng et al., 267-274). Efforts in the field have been put in place to test methods in language-specific corpora and cross-language settings with encouraging results (Stamatatos et al. «Overview»). Javier Blasco and Cristina Ruiz Urbón highlighted the importance of the language and the proper choice of features when applied to Spanish texts. Albeit their study focused on modern Spanish texts extracted from online newspapers and blogs, they still mention the controversy surrounding the *Lazarillo* and noted the peculiarities of dealing with Spanish Golden Age works.

In general terms, modern authorship attribution problems fall in different categories depending on the desired outcome and the corpus. The process of discovering how alike two given texts are and finding their similarities is usually known as plagiarism detection (Stein, Lipka, and Prettenhofer, 63-82; Stein and zu Eissen; Zu Eissen and Benno Stein, 565-569).[41] When the corpus is not available, researchers try to cluster authors, a tech-

---

39.– All these *old* techniques are very well explained in David I. Holmes («Authorship attribution» 87-106).

40.– Such is the case for CUSUM (QSUM) by Andrew Morton and Sidney Michaelson, strongly criticized by David Holmes and Fiona Tweedie (19-47).

41.–  An interesting introduction and approach can be found in Marilyn Randall.

nique that divides up the texts into parts that maintain the same style in order to discern authorship in collaborative works, which makes it possible to show the evolution of an author's style over time (Collins et al. 15-36; Graham, Hirst, and Marthi, 397-415). In some cases, it is even possible to characterize the profile of authors in terms of age, education, etc. by means of their writing (Koppel, Argamon, and Shimoni; Rangel et al.). While these approaches might some day be useful when applied to the *Lazarillo*, unfortunately they are still in their infancy. On the other hand, authorship identification counts with a more solid and dilated history, both in terms of research published and success cases. It is defined generally as the task of determining the unknown author of a given text from a set of candidates whose texts' authorships are generally accepted. Unlike closed-set attribution identification problems where the authors involved are known and the only task remaining is to identify who wrote what, *Lazarillo* turns out to be an open-set problem, where new authors are still being added to the pool of candidates. Open-set problems are considered much more difficult to dilucidate as there is no guarantee that the true author is part of the pool of candidates, especially when its size is small (Koppel, Schler, and Argamon, 83-94). Author verification, the problem of authorship identification with a set of one only candidate, is even more challenging since the task is to determine if the candidate is the author or not (Koppel and Schler). Among the different approaches for authorship identification, some scholars treat the problem as a set of instances of author verification, one per each author in the candidate set (Craig).

According to Hugh Craig, non-traditional authorship attribution studies lay their foundation on the idea of writers being constrained by their own cognitive faculties, resulting in a finite and statistically analyzable set of variation patterns that form their style (Craig). As it appears, authors cannot escape their style, not even when writing in different genres since «much of language production is done by parts of the brain which act in such swift and complex ways that they can be called a true linguistic unconscious.»[42] As tenable as they seem, style markers do not convey the power of conviction that traditional humanities scholars consider sufficient. The black-box, or yes-or-no approaches most computer-based authorship studies follow do not provide the stylistic explanation expected by the experts. Computational approaches to authorship attribution, and thus to authorship identification and verification, are not considered sufficient evidence to state the final truth in the dispute of anonymous texts. However, as we demonstrate in this study, using automatic authorship attribution might help reduce the pool of candidates and contribute evidence to further support a specific possible author or set of authors.

## Materials

One big problem of computational methods is that they usually require the availability of digital editions of the texts, and it has been proven that some of the methods work better when their extensions are at the level of entire books. These kinds of collections exist but they do not usually grant access to the whole text;[43] therefore, in order to carry out

---

42.– Citing from Craig, in relation to Mary Thomas Crane.

43.– CORDE, for example, only allows counting frequencies.

our study, we were forced to collect our own corpus. We decided our corpus to comprise works in a period of 90 years surrounding the publication of the first known edition of the *Lazarillo*. All the major aforementioned candidates for the authorship of the little book are included, as well as some authors who had not been considered previously. The inclusion of these other authors is not coincidental: they add robustness to our method and establish a framework to assess its effectivity. We consider the period from 1499 to 1589 to be comprehensive enough to cover the nuances of all possible publication dates, lifespans and active period of authors. This span is even more generous if we take for granted the genetic-literary analysis by Ocasar (*La atribución*), who allegedly found the first citation to the *Lazarillo* in the early editions of *Coloquios de Palatino y Pinciano*, published in 1550.[44]

Collecting a dataset of the kind described was not an easy task. Some of the works are still in manuscript form and lack normalization, modernization, and digitized text, which makes the task even more monstrous. Digitization of original Spanish Golden Age manuscripts also presented some challenges, which we solved by building and using our own crowdsourcing OCR reviewing tool, *i.e.*, Festos.[45] Object character recognition (OCR) is the process of transforming an image of a text into its digital version readable by both people and machines. We built Festos upon DocumentCloud,[46] a platform for journalists to collaboratively share and annotate documents, and Tesseract (Smith, 629-633), a state-of-the-art OCR tool open sourced by Google. While Tesseract is pluggable (Smith, Antonova, and Lee), it still lacks a good understanding of manuscript typefaces and old Spanish. These limitations were overcome by adding a reviewing tool in Festos that allowed collaborators to correct and proofread the results of the automatic recognition. This reviewing feature sped up the process of getting the digital texts ready as compared to the approach of transcriptions from scratch.

Unfortunately, although some works were already in digital form and others had modern usable editions, due to resources and time constraints we were unable to collect works from all the authors proposed and mentioned in this study as possible fathers of the *Lazarillo*. Pedro de Rhúa and Hernán Núñez de Toledo are among the authors without representation in our list of works, although this might not pose a great burden on our study since they were arguably the weakest of the candidates: not supported ever since they were first proposed in 1955 and 1964, respectively. Friars Juan de Pineda and Juan de Ortega, the first and last candidates to date to be proposed, are the other two authors not present in our corpus. The former has not been backed up yet by any other scholar, the latter does not count with any known work that could be used. The final list of works by authors in the pool of candidates analyzed is detailed below:[47]

- Alfonso de Valdés: *Diálogo de las cosas acaecidas en Roma* (1527), *Diálogo de Mercurio y Carón* (1528)
- Diego Hurtado de Mendoza: *De la Guerra de Granada* (~1573)
- Francisco Cervantes de Salazar: *Crónica de la Nueva España* (1575)

---

44.– Fernando Calero dates these *Coloquios* even earlier, around 1539 (Calero, *Los Coloquios*).

45.– Festos. October 30, 2015. ‹http://festos.cultureplex.ca›.

46.– Document Cloud. October 30, 2015. ‹https://www.documentcloud.org/›.

47.– Dates consigned are publication dates or around the date of death of the author if posthumously published.

- Juan Arce de Otálora: *Coloquios de Palatino y Pinciano* (1550)
- Juan de Valdés: *Consideraciones* (1575), *Diálogo de la Lengua* (1535), *Trataditos* (1545)
- Juan Luis Vives: *Ejercicios de lengua latina (Diálogos)* (1539), *El Alma y la Vida* (1538), *Sobre el socorro de los pobres o Sobre las necesidades humanas* (1525), *Sobre la Concordia y la Discordia* (1529), *Instrucción De La Mujer Cristiana* (1523), *La Sabiduría* (1544), *Las Dimensiones de Europa y del Estado* (1526), *Las Disciplinas* (1531), *Los Deberes del Marido* (1528)
- Lope de Rueda: *Armelina* (~1565), *Auto de Naval y de Abigail* (~1565), *Coloquio de Camila* (~1565), *Coloquio de Tymbria* (~1565), *Discordia* (~1565), *El Deleitoso* (~1565), *Eufemia* (~1565), *Farsa del Sordo* (1549), *Los Desposorios de Moisén* (~1565), *Los Engañados* (1560), *Registro de Representantes* (~1565), *Medora* (~1565), *Prendas del Amor* (~1565)
- Sebastián de Horozco: *La famosa historia de Ruth* (~1570), *Relaciones Históricas Toledanas* (~1570)

Furthermore, we added works from coetaneous authors of *Lazarillo*'s: some with evident connections to the circumstance of the little book (Pedro Mejía, Pérez de Oliva), others with no connection whatsoever (Torquemada, Juan de Malara), and a few minor or discarded attributions (Fernando de Rojas).

- Antonio de Torquemada: *Don Olivante de Laura* (1564)
- Cristóbal de Villalón: *El Crotalón de Christophoro Gnophoso* (1552)
- Gaspar Gil Polo: *Diana enamorada* (1564)
- Gonzalo Argote de Molina: *Discurso sobre la Poesía Castellana* (1575)
- Fadrique de Zúñiga y Sotomayor: *Libro de Cetrería* (1565)
- Fernán Pérez de Oliva: *Diálogo de la Dignidad del Hombre* (1586)
- Fernando de Rojas: *La Celestina* (1499)
- Francisco Delicado: *La Lozana Andaluza* (1528)
- Juan de Malara: *Descripción de la Galera Real del Sermo. Sr. D. Juan de Austria* (~1570)
- Pedro Mejía: *Carlos V* (1530), *Coloquios del Convite* (1547), *Coloquio del Porfiado* (1547), *Coloquio del Sol* (1547), *Dialogo de la Tierra* (1547), *Diálogo de los Médicos* (1547), *Diálogo Natural* (1547), *Silva de Varia Lección* (1540)
- Sebastián Fernández: *Tragedia Policiana* (1547)

The corpus counts a total of 50 works by different authors of different genres, styles, and extensions.[48] Regarding *Lazarillo* itself, we used the edition of the Centro Virtual Cervantes, which is a digital edition based on those published in 1554 in Burgos (Spain) by Juan de Junta, Alcalá de Henares (Spain) by Salzedo, Antwerp (Belgium) by Martín Nucio, and Medina del Campo (Spain) by Mateo and Francisco del Canto. The edition, also collated with the critical works by Alberto Blecua, José M. Caso González, and Francisco Rico (Anónimo ed. Blecua; Anónimo ed. Caso González; Anónimo ed. Rico),

---

48.– Gonzalo Argote de Molina's *Poesías Castellanas* was later discarded as its extension was too short to support any statement about authorship.

marks visually the interpolations that the edition of Alcalá added. There is some contro­versy around deciding whether those additions should be considered as apocryphal, or as coming from the same author and therefore part of the *princeps*. In this context, and aiming to improve the accuracy of our method by only having works written by the same author, we segmented the little book and assigned different anonymous authors to each separate part. For purposes of completeness, we also added the second part, *La segunda parte de Lazarillo de Tormes y de sus fortunas y adversidades* (*Second part of the Lazarillo de Tormes and of his fortunes and adversities*), published in 1555 in Antwerp by the printer Martín Nucio. Digitally edited by Centro Virtual Cervantes, it takes into account the editions by Buenaventura Carlos Aribau (Anónimo ed. Carlos Aribau), and the one by Pedro Manuel Piñero Ramírez (Anónimo ed. Piñero). The final list looks as follows:[49]

- Anonymous +: *La vida de Lazarillo de tormes y de sus fortunas y adversidades* (1554) (with interpolations)
- Anonymous -: *La vida de Lazarillo de tormes y de sus fortunas y adversidades* (1554) (without interpolations)
- Anonymous S: *La segunda parte de Lazarillo de Tormes y de sus fortunas y adversi­dades* (1555)

## Methods

In the presentation of their automated tool (JGAAP) Patrick Juola, John Sofko, and Patrick Brennan stated that «all known human languages can be described as an un­bounded sequence chosen from a finite space of possible events.» These events might range from the different words of a language such as Spanish, to the letters of a specific alphabet, or the different phonemes in the spoken inventory; as such, any written book meets the definition. They also considered that, generally, almost any non-traditional au­thorship attribution analysis —and, thus, author identification— can be seen as a three-phases pipeline, each of which must be tailored to the specific needs of the corpus and task at hand (Juola, Sofko, and Brennan). We adopted their framework for its broad and comprehensive view and redefined the steps for our purposes. The first one, canonical­ization,[50] is the process of standardizing the events in the text in order to reduce the complexity and thus the number of different symbols and words to handle. The rules we followed for regularizing the spelling of old Spanish were borrowed from Ocasar's sys­tem in his edition of the *Coloquios de Palatino y Pinciano* by Arce de Otálora (Ocasar, *La atribución*), to which we added some of our own. Specifically, we removed margin anno­tations and footnotes; removed page headers, footers, and numbers; removed any Latin or Greek citations; joined split words; removed spurious characters; removed duplicated punctuation marks; converted all possible hyphens into one; removed numbers in text as they usually add little to the style; expanded abbreviations such as «Đ» into the ca­nonical form «DE;» and removed starting and ending marks of chapters, volumes, parts,

---

49.– Unfortunately, the interpolations are not long enough to be included in the authorship attribution study. Possible workarounds for this issue are discussed in the conclusions and further research of this study.

50.– «Canonicization» in the original text.

scenes, and books. For plays, we also removed names of speakers. Then the event set had to be determined, which includes the partitioning of the works in the corpus into non overlapping events, such as paragraphs, sentences or words. The last step was the application of different kinds of statistical inferences to said events, from basic frequencies and distance-based measures to machine learning and pattern-based techniques. The specific features to be extracted depend on the statistical analysis to be carried out. This process can be seen as a transformation of the text into numbers, an ultimate quantification that produces vectors from stories attending to a variety of criteria: a corpus is now transformed into a more general and abstract dataset. The main goal of any feature extraction step is to maximize the discriminative power of the feature set selected, that may contain different kinds of features. Efstathios Stamatatos classifies the features in 5 groups, according to their nature and role in the text, and each requiring different mechanisms for their obtention (see table 2). Lexical and character features are historically the first ones to be used, and deal with the text at the word and letter levels, respectively. Frequency distributions of words or characters (bag of words), or ordered sets of them of different lengths (n-grams) are among the most used lexical features and the ones that provide best results. Although they are very useful since they can be applied regardless of the language, the extraction of lexical features might require the use of advanced techniques from natural language processing in order to segment the text into sentences or words —tokenizers, stemmers, and lemmatizers may come in handy. Extraction of semantic and syntactic features involves an even more sophisticated analysis of the texts, as it uses layers of knowledge that are not revealed in the text itself. These abstract constructs such as parts of speech, polysemy, or phrase structure, are related to a specific role of parts of the text. In practice, the extraction of this kind of features can be thought of as a two-step process: first, the text is transformed according to the function of its parts, and second, the same mechanism of counting the lexical features can then be used.

Furthermore, Stamatatos also makes a distinction according to how the different methods of attribution treat the corpus. Profile-based approaches operate on a per-author basis, concatenating all texts by the same author and extracting the features cumulatively, ignoring in fact the possible existence of differences amongst their texts. Instance-based methods, on the other hand, treat each text individually and try to produce most accurate attribution models by considering the individual contributions that each of the texts makes to the authorial style. Generally, as a manner to artificially increase the number of texts available in the corpus, chunking the works into parts of equal sizes in terms of number of paragraphs, sentences, or words is a widely employed technique. Finally, there is a third approach that would combine both profile and instance-based methods. Regardless of the technique of attribution used, the selection of features and their size or dimensionality still remains a rather arbitrary and domain specific task.

Table 2: Summary of features by category following Stamatatos' classification and adding some from Argamon and Juola's overview (Argamon and Juola, «Overview»).

| Category | Features |
|---|---|
| Lexical | Token-based (word length, sentence length, etc.) |
| | Vocabulary richness |
| | Word frequencies |
| | Word n-grams |
| | Errors |
| | Function words |
| | Pronouns |
| | Modal verbs |
| | Contractions/abbreviations |
| Character | Character types (letters, digits, etc.) |
| | Character n-grams (fixed-length) |
| | Character n-grams (variable-length) |
| | Compression methods |
| | Punctuation |
| | Suffixes |
| Syntactic | Part-of-Speech |
| | Chunks |
| | Sentence and phrase structure |
| | Rewrite rules frequencies |
| Semantic | Synonyms |
| | Semantic dependencies |
| | Semantic parser |
| | Named entity types |
| | Polysemy / specificity |
| Application-specific | Structural |
| | Content-specific |
| | Language-specific |

## Comprehension and Compression

When faced with many features, dimensionality reduction and feature selection techniques can be applied (Forman), although they might fail to capture authors' styles and therefore result in features too genre- or topic-dependent (Brank et al.). John Burrows, after experimenting with techniques based on multivariate analysis to reduce the dimensionality of the feature space, came up with an approach that fits perfectly in Juola's broad definition of an authorship attribution method: the 'Delta' method (Burrows, «Delta» 267-287; «Attribution and Beyond»). From a frequency distribution of the 150 most frequent words in a corpus, the method starts by estimating the mean frequency of the word and its variance, the so called z-distribution. Burrows' 'Delta' (which he insisted it to be named 'Δ' where possible, although his claims were unheard) is then built as «the mean of the absolute differences between the z-scores for a set of word-variables in a given text-group and the z-scores for the same set of word-variables in a target text.» Which means that the smaller the Delta, the more similar the texts are. This profile-based method turned out to be the most robust single measure and it is now used as a baseline for other methods since it usually produces useful results across genres and languages. Some improvements have been proposed based on explanations of the underlying mathematics involved, but Burrows' 'Delta' has proven over and over to perform better than its modifications despite lacking a solid theoretical background (Stein and Argamon, «A mathematical explanation» 207-209; Rybicki and Eder, «Deeper Delta» fqr031).

Table 3: Best Deltas for our corpus. Each row shows a different setting for culling and most frequent words, the best performing Delta in each case, and the difference of means as defined by Jannidis as a proxy for best measure.

| Most frequent words | Culling | Delta | Difference of means |
| --- | --- | --- | --- |
| 100 | 50% | Eders Delta | 1.50 |
| 100 | 70% | Eders Delta | 1.50 |
| 2500 | 0% | Cosine | 1.49 |
| 100 | 90% | Euclidean | 1.49 |
| 500 | 90% | Eders Delta | 1.48 |
| 1000 | 90% | Eders Delta | 1.48 |
| 2500 | 90% | Eders Delta | 1.48 |
| 500 | 70% | Eders Delta | 1.46 |
| 1000 | 70% | Eders Delta | 1.46 |
| 2500 | 70% | Eders Delta | 1.46 |
| 500 | 50% | Eders Delta | 1.46 |
| 100 | 0% | Canberra | 1.45 |
| 100 | 30% | Canberra | 1.45 |
| 1000 | 50% | Eders Delta | 1.44 |
| 2500 | 50% | Eders Delta | 1.44 |
| 2500 | 30% | Cosine | 1.42 |
| 500 | 30% | Eders Delta | 1.41 |
| 1000 | 0% | Cosine | 1.41 |
| 1000 | 30% | Eders Delta | 1.40 |
| 500 | 0% | Eders Delta | 1.39 |

Fotis Jannidis recently proposed a framework based on a simple difference of means to evaluate and assess the 'Delta' method and its variations. The measure «showed the best correlation with the clustering error measure» when doing ingroup and outgroup comparisons —ingroup refers to distances between texts written by the same author, and outgroup by different authors. The larger the difference, the better the measure performs. They also published the code used to carry out their analysis —a practice that we believe should become more common—, which we used with slight modifications over our corpus of the *Lazarillo* ( Jannidis, et al.; Evert et al.). We executed several runs changing the number of most frequent words to consider (150, 500, 1000, 2500), and also applied different *culling* factors (30%, 50%, 70%, 90%) based on David Hoover's extensive analysis and variations over the original 'Delta' method.[51] We obtained that Maciej Eder's variation, a variant derived from the Canberra measure of similarity (Rybicki and Eder, *Deeper Delta*), performed sensibly better than baseline and than more sophisticated Deltas such as cosine-based or simpler ones such as the Euclidean (see table 3 for a summary of the executions). This might be explained by the fact that Eder's Delta seems to provide better results for highly inflected languages, and although only tested for French, it might work as well for Spanish (Eder and Rybicki, «Birds of a feather» fqs036; Eder, «Does size matter?» 132-135).

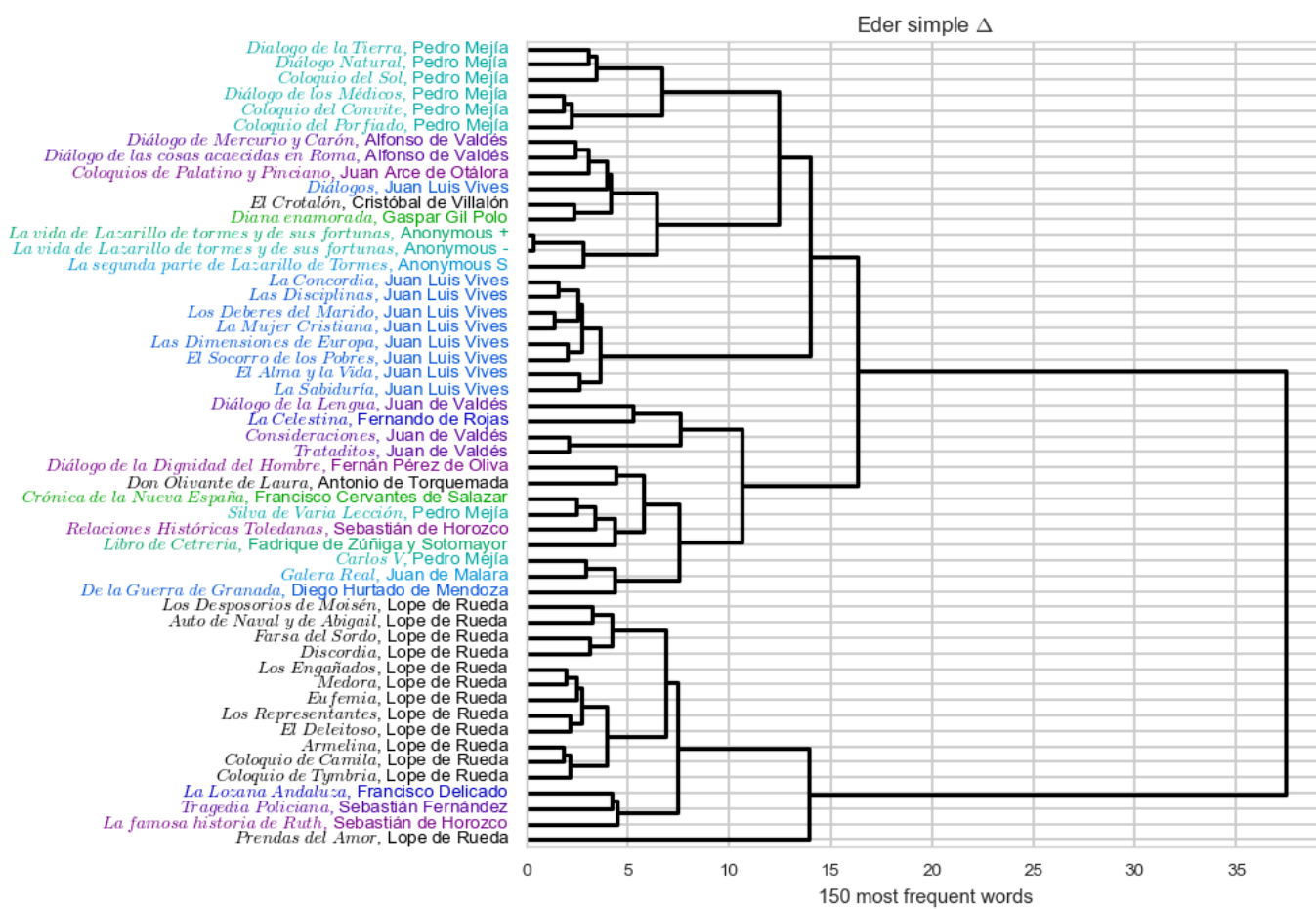

Figure 2: Dendrograms for Eder simple Delta. Jannidis' algorithm performs an arborean grouping by similarity measured by the chosen Delta distance. Eder's simple Delta is calculated with 0% of culling and for the 150 most frequent words. Same authors are assigned the same color.

51.– Culling is the percentage of documents a word must appear in to be retained in the corpus (Hoover, «Delta prime?» 477-495; «Testing Burrows's delta» 453-475).

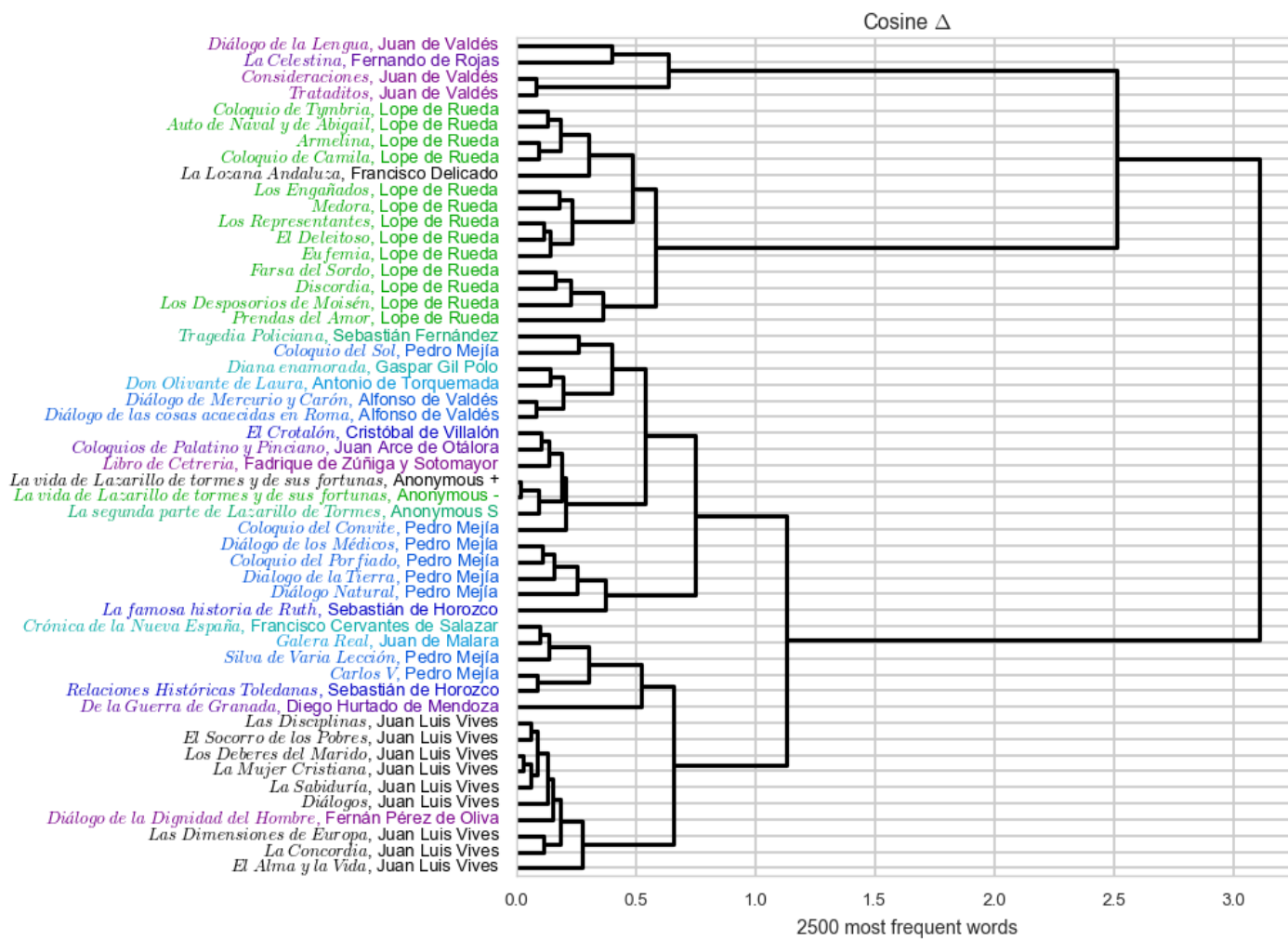

Figure 3: Dendrograms for Cosine Delta. Cosine Delta is calculated for the 2500 most frequent words and with 0% of culling. Same authors are assigned the same color.

Figures 2 and 3 show dendrograms that put into a hierarchy the works in our corpus by means of the Delta measure. Works in leaves with the same parent are closer to each other than to those works in leaves accessible only by traversing the tree. Following the arborean structure Jannidis' algorithm produces, it is easy to see how some of the candidates just stay out of the branch that reaches the *Lazarillo*. The method consistently leaves authors such as Fernán Pérez de Oliva, Fernando de Rojas, Francisco Delicado, or Juan de Malara far from our wanted anonymous author. As a first approach it goes with our intuitions as they were *impostors* in our corpus. However, authors with strong candidacies are also grouped differently than the *Lazarillo*, as it is the case of Juan Luis Vives, Diego Hurtado de Mendoza, Juan Valdés or even Lope de Rueda. Among the ones more closely related to the author of the *Lazarillo*, with or without *interpolaciones*, we find Juan Arce de Otálora and Alfonso de Valdés, but unexpected authors such as Cristóbal de Villalón or Pedro Mejía. The second part of the adventure of Lázaro de Tormes is also placed together with the first two. According to the inventor of the Delta method, this result suggests that we should focus on these last group of authors and reinterpret the authorship of the little book as a closed-set problem. In relation to the dependence of the method on probabilities, Hoover observed that with specific cases of corpora «false attributions are a serious possibility» (Hoover, *Testing*), and Burrows also noted that «the system for distinguishing between insiders and outsiders is not foolproof» (Burrows, 'Delta'). The problem worsens when dealing with either texts of different lengths, or with a different number of texts by author —the class imbalance problem—, as it is our case.[52] Therefore, before making a hasty decision, we had better explore other methods for authorship to further support this initial findings.

Another set of distance-based methods borrows some concepts from the principles that make regular file compression applications work. Their functioning is similar to other probabilistic distance measures such as those based on Markov models (Khmelev and Tweedie, «Using Markov» 299-307; Kukushkina, Polikarpov, and Khmelev, «Using literal» 172-84),[53] but avoids the combinatorial explosion problem when facing huge vocabulary sizes. In general, as compression methods are usually, but not necessarily, profile-based approaches, the first step is to concatenate all the works by an author to later compress them into individual files. The anonymous text is then added to the concatenated files and they are compressed again. The bitwise difference between the concatenated text files with and without the anonymous text is a measure of the similarity of both texts and can be used as a proxy for authorship attribution. Technically, from an information theory perspective, compression methods calculate the cross-entropy or Kullback-Leibler divergence of the texts as a measure of closeness.[54] Fortunately, despite the mathematical complexity of this measure, the underlying idea is so easy to grasp that we could bypass the formulas by using virtually any compression tool available. In this scenario, the RAR compression format has shown to outperform any other, specially dictionary-based compression such as LZMA or GZIP (Khmelev and Teahan, «A repetition» 104-110; Marton, Wu, and Hellerstein, 300-314).

In this context we used a technique implemented in 2005 by Rudi Cilibrasi and Paul Vitanyi around BZIP2, another very popular, free, and open source compression format (45). Let $C(x)$ be the bitwise size of the compression of a text file x, and let denote concatenation of files x and y as x+y. Cilibrasi and Vitanyi built upon the concept of Kolmogorov complexity and defined their normalized compression distance (NCD) between the files x and y as follows:[55]

$$\text{NCD}(x, y) = \frac{(C(x + y) - \min(C(x),C(y))}{(\max(C(x),C(y))} \qquad (1)$$

52.– Other distance-based methods, such as Common n-Grams (CNG) approach by Keselj et al. are also known to perform poorly under such circumstances (Kešelj et al.; Stamatatos, «Author identification» 237-241).

53.– A good introduction to the topic with applications and examples can be found in Ming Li and Paul Vitnyi.

54.– Joula referred to the method as the «linguistic cross-entropy» («What can we do» 1; «Cross-entropy» 141-149).

55.– Defined as the «length of the smallest computer program that converts one string into another... authorship can [therefore] be assigned to the training document that would require the least 'work' to convert to the test document» (Juola, *Authorship*).

Broadly defined, Kolmogorov complexity «is a measure of the computational resources needed to specify an object» in an universal descriptive language (Burgin). In our case the object is a text, understood as a digital string of characters, and the computational resources can be specified as the length of the shortest computer program —written in any prefixed programing language— able to produce such an output.

The researchers reported excellent results for Russian texts, and even for their machine-translated English versions, as well as in other fields such as music and genomics. Other have reported that the technique might be noise-robust, that is resistant to noise (Cebrián, Alfonseca, and Ortega, 1895-900), which despite our efforts manually curating is still a reality in our corpus. Inspired by the alleged efficacy we applied Cilibrasi and Vitanyi's method virtually unchanged by using a library they released and containerized for others to use: CompLearn Toolkit (Cilibrasi, *CompLearn*). Once the distances between each pair of texts (or concatenated texts per author) are calculated, NCD provides us with a tool to cluster them by their similarity and represent them using a hierarchy. The result is an unrooted binary tree in which leaves in the same level have closer small distances. Figures 4 and 5 show our results for two different runs: first using an instance-based approach, and second a profile-based one. When texts are grouped by author (figure 4),[56] NCD shows that the first and second part of the *Lazarillo* cluster pretty closely together, followed by Fernán Pérez de Oliva, Francisco Cervantes de Salazar, and Francisco Delicado. Out of these last three, two are part of the impostors section of our corpus and the third, Cervantes de Salazar, although supported by Madrigal using computational means, was later rejected. In the next level we find a mix of impostors and genuine candidates: Sebastián Fernández, Hurtado de Mendoza (linked to Juan de Malara), Gaspar Gil Polo, Fernando de Rojas, and Alfonso de Valdés. The furthest positions belong to Juan Arce de Otálora and Pedro Mejía. By all means, these results practically contradict Delta's. We believe that the incomprehensive groupings performed in the clustering provided by the NCD tool must be sensible to the class imbalance problem, as there seem to be a slight relation between the length of the concatenated texts and the closeness at which authors are clustered. On the other hand, results for the instance-based approach (see figure 5)[57]

56.– Anonymous + as «A», Anonymous S as «AS», Juan Arce de Otálora as «JAO», Francisco Cervantes de Salazar as «FCS», Francisco Delicado as «FD», Sebastián Fernández as «SF», Gaspar Gil Polo as «GGP», Sebastián de Horozco as «SH», Diego Hurtado de Mendoza as «DHM», Juan de Malara as «JM», Pedro Mejía as «PM», Fernán Pérez de Oliva as «FPO», Fernando de Rojas as «FR», Lope de Rueda as «LR», Antonio de Torquemada as «AT», Alfonso de Valdés as «AV», Juan de Valdés as «JV», Cristóbal de Villalón as «CV», Juan Luis Vives as «JLV», and Fadrique de Zúñiga y Sotomayor as «FZS.»

57.– *La vida de Lazarillo de tormes y de sus fortunas y adversidades* as «A,Lazarillo», *La segunda parte de Lazarillo de Tormes* as «AS,Lazarillo», *Coloquios de Palatino y Pinciano* as «JAO,Coloquios», *Crónica de la Nueva España* as «FCS,Crónica», *La Lozana Andaluza* as «FD,Lozana», *Tragedia Policiana* as «SF,Tragedia», *Diana enamorada* as «GGP,Diana», *La famosa historia de Ruth* as «SH,Ruth», *Relaciones Históricas Toledanas* as «SH,Relaciones», *De la Guerra de Granada* as «DHM,Guerra», as «JM,Galera», *Carlos V* as «PM,Carlos», *Coloquio del Convite* as «PM,Convite», *Coloquio del Porfiado* as «PM,Porfiado», *Coloquio del Sol* as «PM,Sol», *Dialogo de la Tierra* as «PM,Tierra», *Diálogo de los Médicos* as «PM,Médicos», *Diálogo Natural* as «PM,Natural», *Silva de Varia Lección* as «PM,Silva», *Diálogo de la Dignidad del Hombre* as «FPO,Dignidad», *La Celestina* as «FR,Celestina», *Armelina* as «LR,Armelina», *Auto de Naval y de Abigail* as «LR,Naval», *Coloquio de Camila* as «LR,Camila», *Coloquio de Tymbria* as «LR,Tymbria», *Discordia* as «LR,Discordia», *El Deleitoso* as «LR,Deleitoso», *Eufemia* as «LR,Eufemia», *Farsa del Sordo* as «LR,Sordo», *Los Desposorios de Moisén* as «LR,Moisén», *Los Engañados* as «LR,Engañados», *Los Representantes* as «LR,Representantes», *Medora* as «LR,Medora», *Prendas del Amor* as «LR,Amor», *Don Olivante de Laura* as «AT,Olivante», *Diálogo de las cosas acaecidas en Roma* as «AV,Roma», *Diálogo de Mercurio y Carón* as «AV,Mercurio», *Consideraciones* as «JV,Consideraciones», *Diálogo de la Lengua* as «JV,Lengua», *Trataditos* as «JV,Tratadi-make more sense as works belonging to Lope de Rueda are clustered together, as it happens to a lesser extend to those by Juan Luis Vives and those by Pedro Mejía. This provides a more solid foundation to interpret the rest of the tree as the method seems to be capturing stylistic similarities rather than text lengths. The *Lazarillo*, with and without interpolations, is first placed close to *La Sabiduría* (*The Wisdom*) by Vives, and in a second level to his *Las Dimensiones de Europa* (*Dimensions of Europe*), the second part of the little book, and to *Diálogo de las cosas acaecidas en Roma* by Alfonso de Valdés. Further levels show heterogeneity of authors and their works with some smaller clusters. Although the instance-based approach shows some signs of coherence, it still lacks credibility. In order to further test the method we decided to implement our own approach with more solid compression formats other than BZIP2, specially PPM and RAR.

Markov-based methods have been reported to produce good results in text. Prediction by partial matching (PPM) is one of that kind: a probabilistic compression technique —achieving lossless compression in text— that creates a model with the likelihood of each letter appearing after each other. Unfortunately, although Cilibrasi and Vitanyi claimed that their tool was able to work with other compression formats, we were unable to put PPM to work with the NCD tool, so we built our own NCD implementation in Python based on the Debian package ppmd by Dmitry Shkarin and added support for RAR by Alexander Roshal (Shkarin, 202-211). For representing the results, and due to the lack of the NCD semi-automated output that included the result of the clustering process, we calculated the correlation matrix for all pairs of instances and profiles and plotted them into a heatmap and a dendrogram (clustermap).[58] The color map indicates closer similarities with darker colors whereas light colors denote more distance.

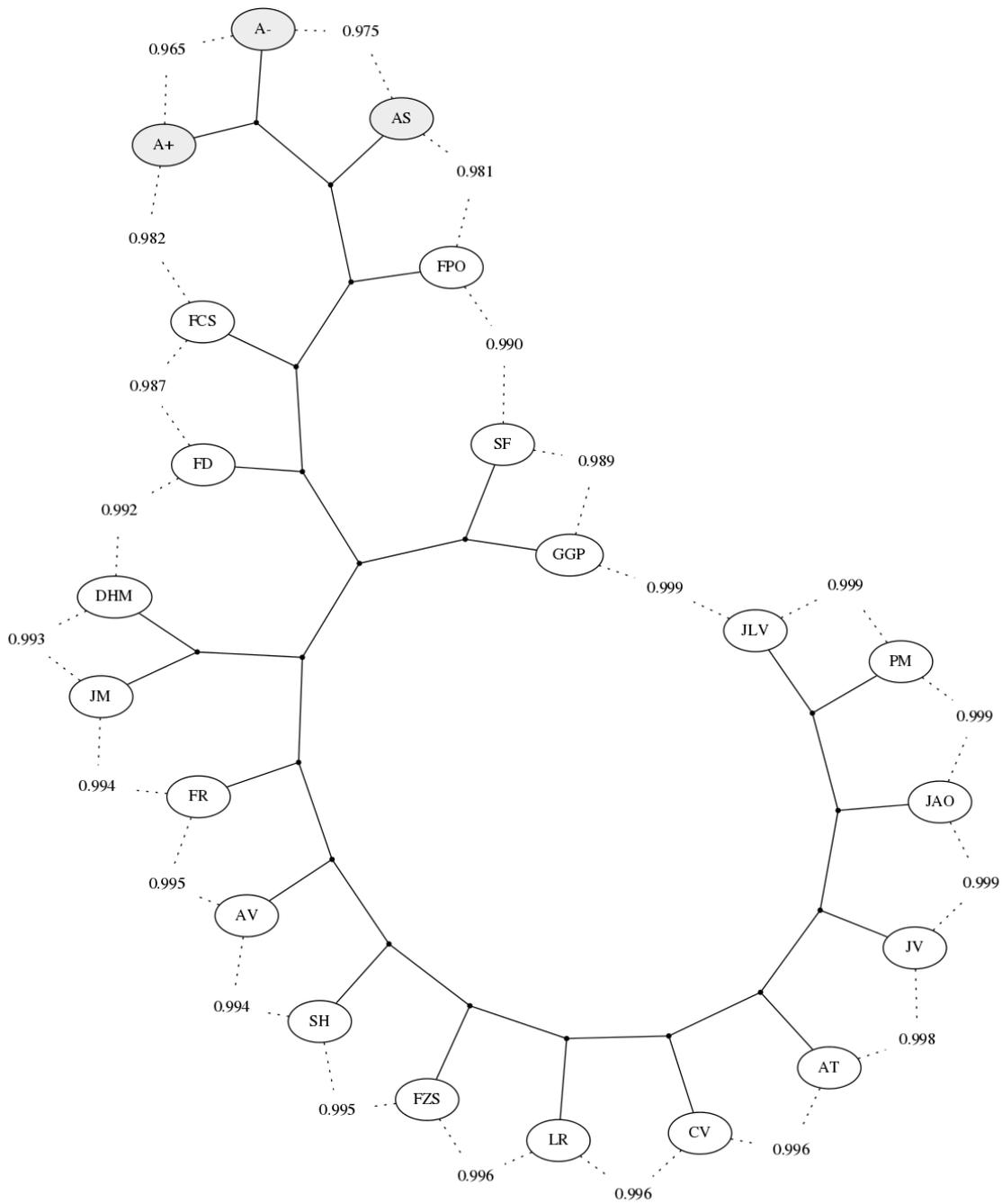

Figure 4: Unrooted binary tree from a matrix of normalized compression distances (profile-based). Some of the distances are included as returned by the NCD tool. Author names have been shortened to avoid overlapping in the graph.

Figure 5: Unrooted binary tree from a matrix of normalized compression distances (instance-based). Leaf labels follow same author codes used in figure 4, whereas work titles are shortened but recognizable.

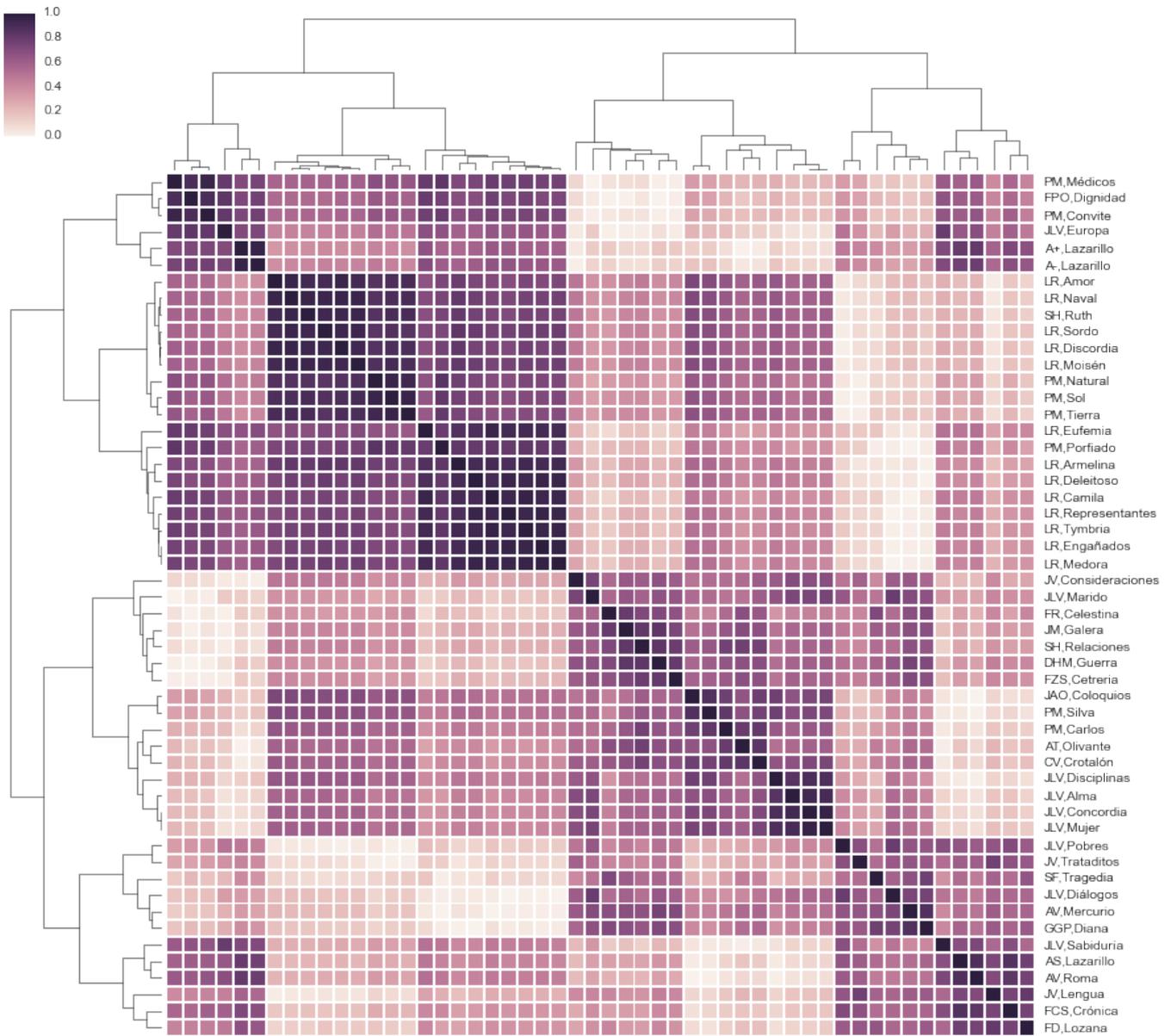

Figure 6a: Instance-based heatmaps and dendrograms for RAR compression format

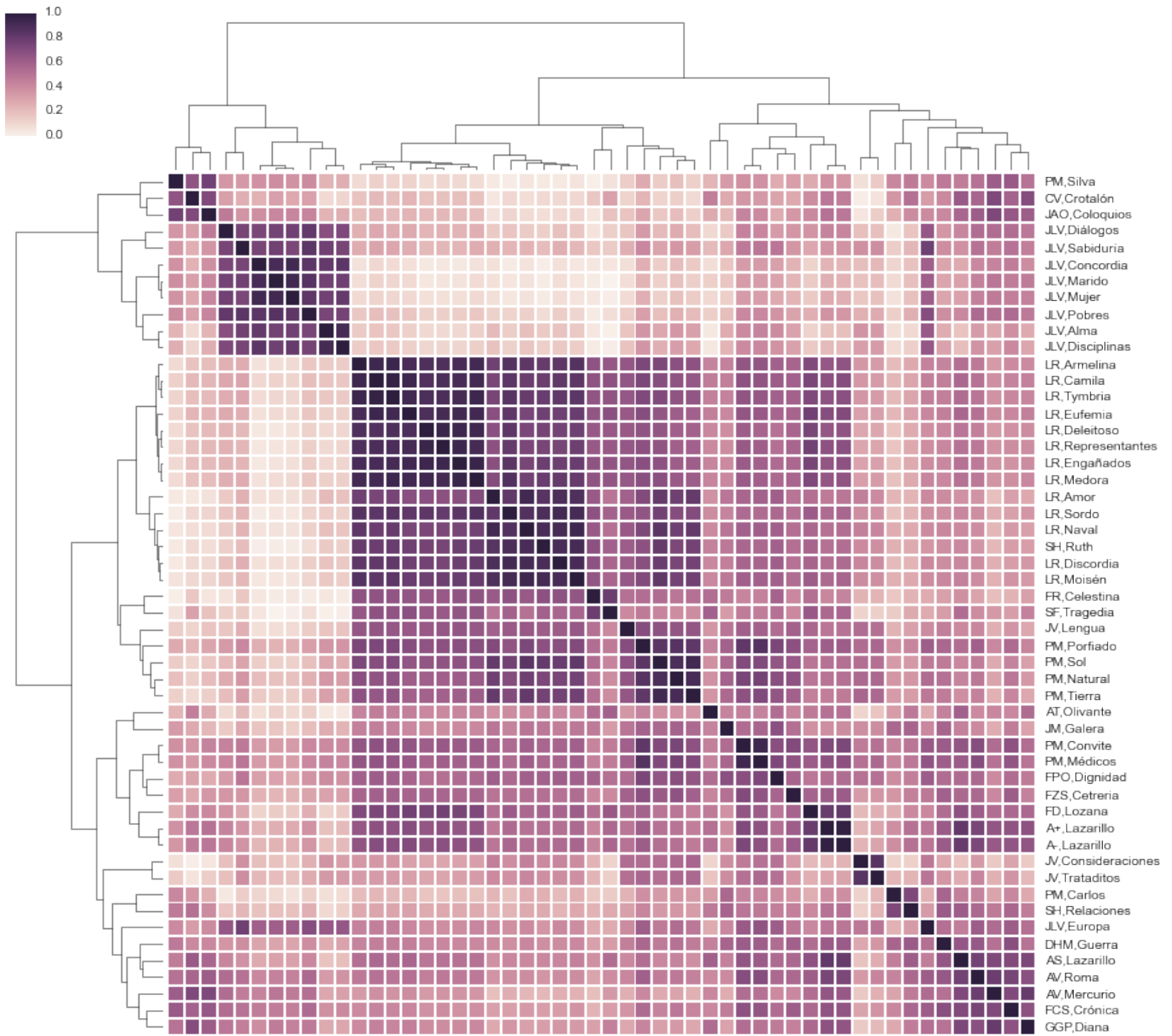

Figure 6b: Instance-based heatmaps and dendrograms for PPM compression format

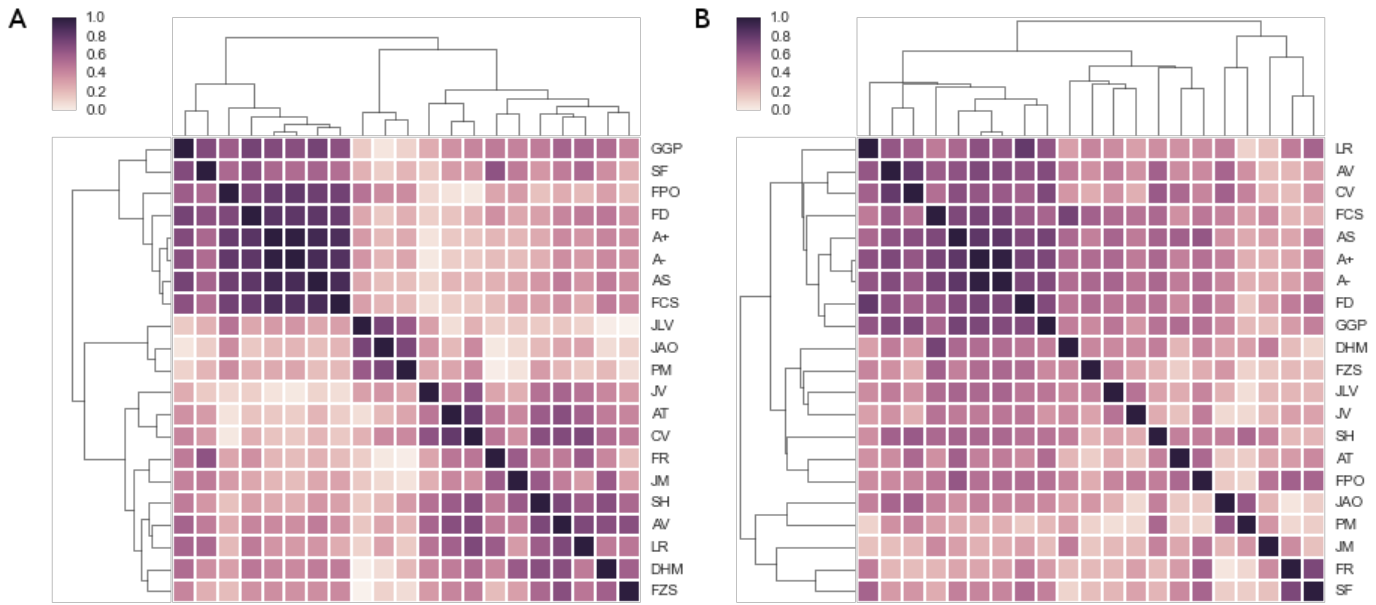

Figure 7: Profile-based plots for RAR and PPM compression formats. Heatmap and dendrogram for the profile-based approach using our own implementation of NCD combined with the (A) RAR and (B) PPM compression formats.

As seen in figure 7 no significant difference is noted between BZIP2, PPM, and RAR for the profile-based approach:[59] the three methods report different groupings of authors. Despite some clusters, otherwise irrelevant for our current study, that seem to remain together such as Juan Arce de Otálora's *Coloquios de Palatino y Pinciano* and Pedro Mejía's *Silva de varia lección* (*A Miscellany of Several Lessons*), the only ones in common among the different methods are Francisco Delicado and Gaspar Gil Polo. In this regard it should be recalled that Navarro Durán insisted in the influx that works such as Francisco Delicado's *La Lozana Andaluza* —that accounts for a reference to a such «Lazarillo»— had in the little book. While huge in Italy, the diffusion of *La Lozana Andaluza* in Spain was small compared to other alleged readings —according to Navarro Durán— made by the author of the *Lazarillo* such as *La Celestina* by Fernando de Rojas. In his critique against Navarro Durán's proposal in favour of Alfonso de Valdés (Pérez Venzalá, El Lazarillo), Pérez Venzalá grants that Delicados' work was still influential to the posterior 16th-century Spanish prose, but that the fact by itself is not enough to argue in favour of the candidacy of the Andalusian. Others were even less inclined to recognize such influence in the *Lazarillo* (Wardropper, 88; Carilla, 97-116). Regarding Gaspar Gil Polo there is simply no mention whatsoever of his implication in the little book; the notary wrote mostly pastoral romances of a very cult tone.

The instance-based approach, however, shows more coherent results. Overall, PPM and RAR clusters are more consistent between approaches, same authors tend to be found regardless. The groups for PPM and RAR share several pairs of (author, work) closely related to the *Lazarillo* and its second part: *Diálogo de los Médicos* (*Dialogue of the Physicians*) and *Coloquio del Convite* (*Colloquia of the banquet*) by Pedro Mejía, and Fernán

---

59.– BZIP2 and RAR NCD performed almost identically.

Pérez de Oliva's *Diálogo de la Dignidad del Hombre* (*Dialogue of the Dignity of Men*). Unfortunately, the only evidence we have about both the Sevillian humanist Mejía and the translator of the classics Pérez de Oliva in relation to the *Lazarillo* is that they moved in the same circles that surrounded the true author.

Notwithstanding, there are some facts that need to be accounted for in relation to the personal information of Pedro Mejía. Born in Seville in 1497, spent his days as a student in Salamanca and cultivated the friendship of important figures of his time such as Erasmus of Rotterdam and Juan Luis Vives. Mejía held several relevant positions in Seville before becoming the official chronicler of Charles V in 1548 after Antonio de Guevara's death. By then he had already written his hugely popular *Silva de varia lección*, that was translated to several languages and circulated all over Europe with tremendous success. Other works of his were published in Antwerp at his death in 1551. In the literary aspect, some lexical coincidences with the little book have been reported, such as «fasta» or «home,»[60] although their effect might have been minimized in our corpus due to the normalization process applied. Nevertheless, more inscrutable aspects of his writing style might have been brought into relevance by our analysis. Pedro Mejía seems to be a figure that demands a more thorough analysis.

## Decomposing the *Lazarillo*

Despite the turn to the abstractness, there is enough evidence to sustain that more convoluted and less intuitive features carry more discriminative power than arbitrary distributions of words or expressions or over simplistic reductions of writing styles to single measures or distances (Argamon and Juola, «*Overview*»). Simple relative or standardized frequencies of words, although presenting both advantages and disadvantages (Forsyth and Holmes, 163-174), are usually preferred in traditional studies since they convey understandable meaning otherwise hidden in unfathomable statistical variables. In their 1988 pioneer study, before embracing his 'Delta' method, John Burrows and Anthony Hassall solved a disputed authorship based on what they called eigenvectors of the correlation matrix from different authors' function words (usually the most common in a language; also called «stop words» in information retrieval studies) (Burrows and Hassall, 427-453). Posterior studies confirmed the separation ability of the «eigenanalysis» under a variety of cases, both in terms of the features used (function words, syntactic tags, etc.) and the works to analyze. The technique was later renamed to its proper and original statistical name: Principal Components Analysis (PCA) (Smith, «Attribution by statistics» 233-251; «The authorship» 508; «Edmund Ironside» 202-205; Binongo, «Incongruity» 477-511; «Joaquin's» 267-279; Binongo and Smith, «The application» 445-466). As a general technique for multivariate analysis, the goal of PCA is to reduce the dimensionality of the vector of features, *i.e.*, transform a frequency distribution of the most frequent 300 words of a text into a pair of values, by summarizing them into new uncorrelated vectors, the so-called principal components, that account for the maximal amount of

60.– Concordance of «home» for «hombre» («man») documented by Diego Clemencín (58), and «fasta» for «hasta» («until») by Rodríguez López-Vázquez (*El 'Tractado'*).

information that can be attributed to them (variance). Principal components are sorted by the power to retain the variation of the original vectors, and as such, the first two or three components are usually used, as they can also be represented graphically, avoiding the hassle of understanding huge correlation matrices.

Table 4: Winner feature sets as extracted from different competitions on authorship problems

| Features | Description |
| --- | --- |
| stopwords | Distribution of functions words |
| bow | Distribution of the 300 most common words (bag of words) |
| cng | Distribution of the 3000 most common character 3-grams (Kešelj et al.; Kourtis and Stamatatos) |
| lexical punctuation lexical + punctuation | Average sentence length, sentence length variation, sentence lexical diversity,[61*] and distribution of punctuation signs |
| pos | Distribution of the 30 most common parts of speech |
| words n-grams | Term frequency-inverse document frequency (tf-idf)[61**] for a maximum of 1000 word bi- and tri-grams |
| characters n-grams | Term frequency-inverse document frequency for a maximum of 1000 character n-grams of length between 2 and 4 |
| total | All above features combined into one single vector |

We used Burrows' approach and ran a PCA on our corpus mimicking his same setup. As the text of the *Lazarillo* itself is not considered very long when compared to other candidates' works, the segmentation of the works in chunks of 150, 300, or 500 words did not have much effect in the results.[61] Nor did the inclusion of the *interpolaciones*, nor the number of stop words used; we tried with 25, 50, 150, and 300 with similar outcomes. As shown in figure 8, in the best case we achieved components that accounted in average for less than 10% of the variance. Nevertheless, basic PCA still remains a useful first step in order to get a glance of a dataset. It is easy to identify visually how some of our random candidates in the corpus stand out as the representation of their chunks in the general plot are easily distinguishable from those of the little book. The clearer the clusters, the less the authors have in common. As such, the authors who exhibit a more similar use of function words are Juan Arce de Otálora, Gaspar Gil Polo, Alfonso de Valdés, Cristóbal

---

61.– The list of function words was extracted from the Python package for natural language processing NLTK, which includes the lists of stopwords for 11 languages compiled by Martin F. Porter in his work with stemmers (130-137; Bird, Klein, and Loper). All the analysis and rendering in this study were made in Python with the use of several packages: numpy, scipy, scikit-learn, Pandas, matplotlib, IPython, and Jupyter are among the most important ones (Pedregosa et al.; Oliphant; McKinney; Jones, Oliphant, and Peterson; Perez; Ragan-Kelley et al.).

61*.– Vocabulary richness, defined as the ratio between the number of different words and the number of total words per sentence.

61**.– The tf-idf measure aims to reflect how important a word is to a text in a given corpus. It was introduced by Gerard Salton and Michael McGill as the ratio of two previous measures, the frequency of a word (tf) and the frequency of that word in the whole corpus (idf). It has been very widely used and applied in information retrieval studies ever since (Salton and McGill).

de Villalón, and to a lesser extent, Pedro Mejía and Juan Luis Vives, names that are already mentioned in our previous analysis. The rest form more or less easily identifiable clusters, thus being the use of stop words different between them. We found no difference for the second part of the *Lazarillo* or taking out the interpolations.

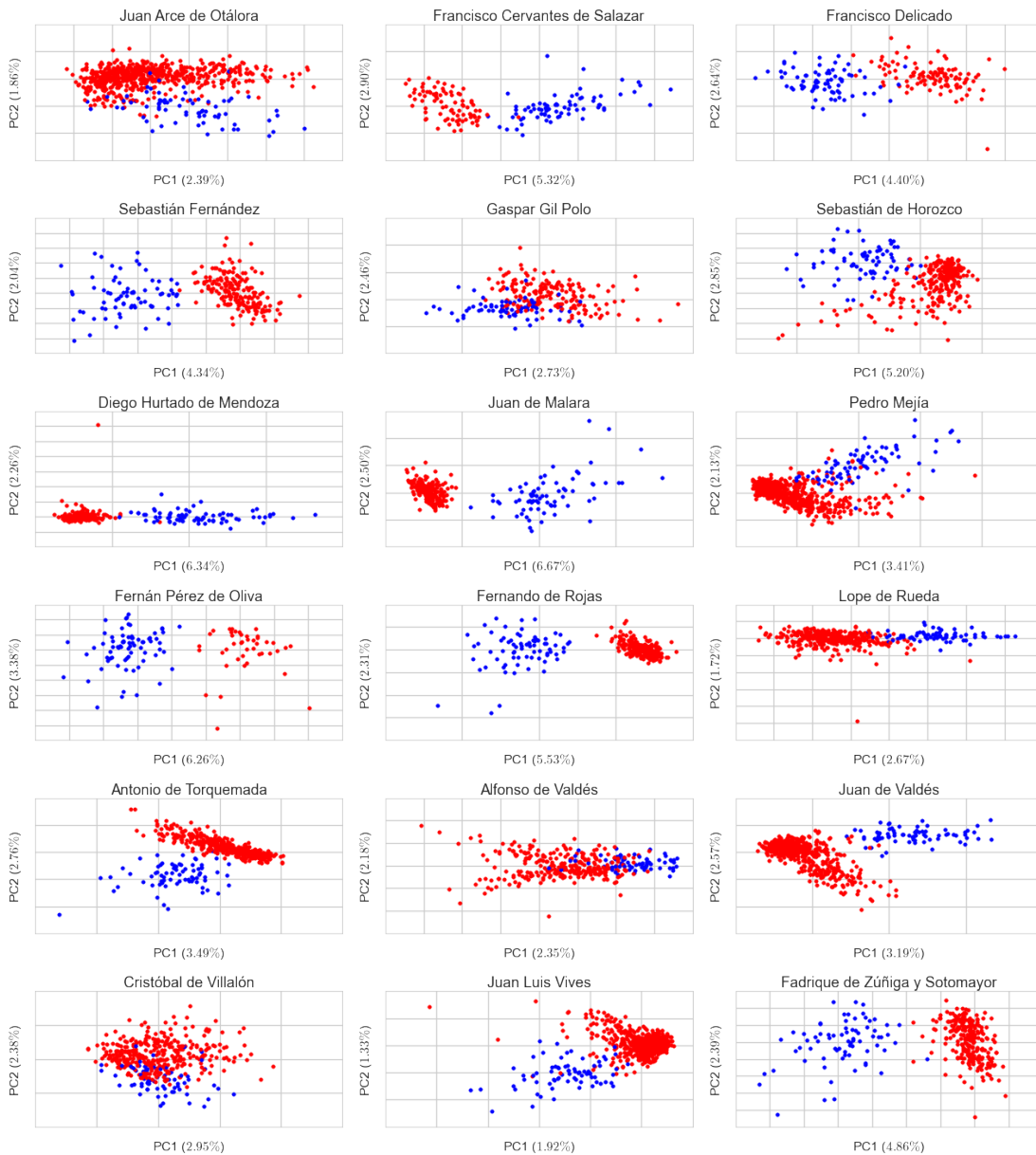

Figure 8: PCA of function words in our corpus. Charts represent the 2 principal components vectors of the frequency distribution of 300 stop words in the *Lazarillo* (blue) and the combined works of each of the possible candidates in the corpus (red). Only 600 random chunks of 300 words are represented, although all were taken into account during the analysis. Variance is shown as axes labels.

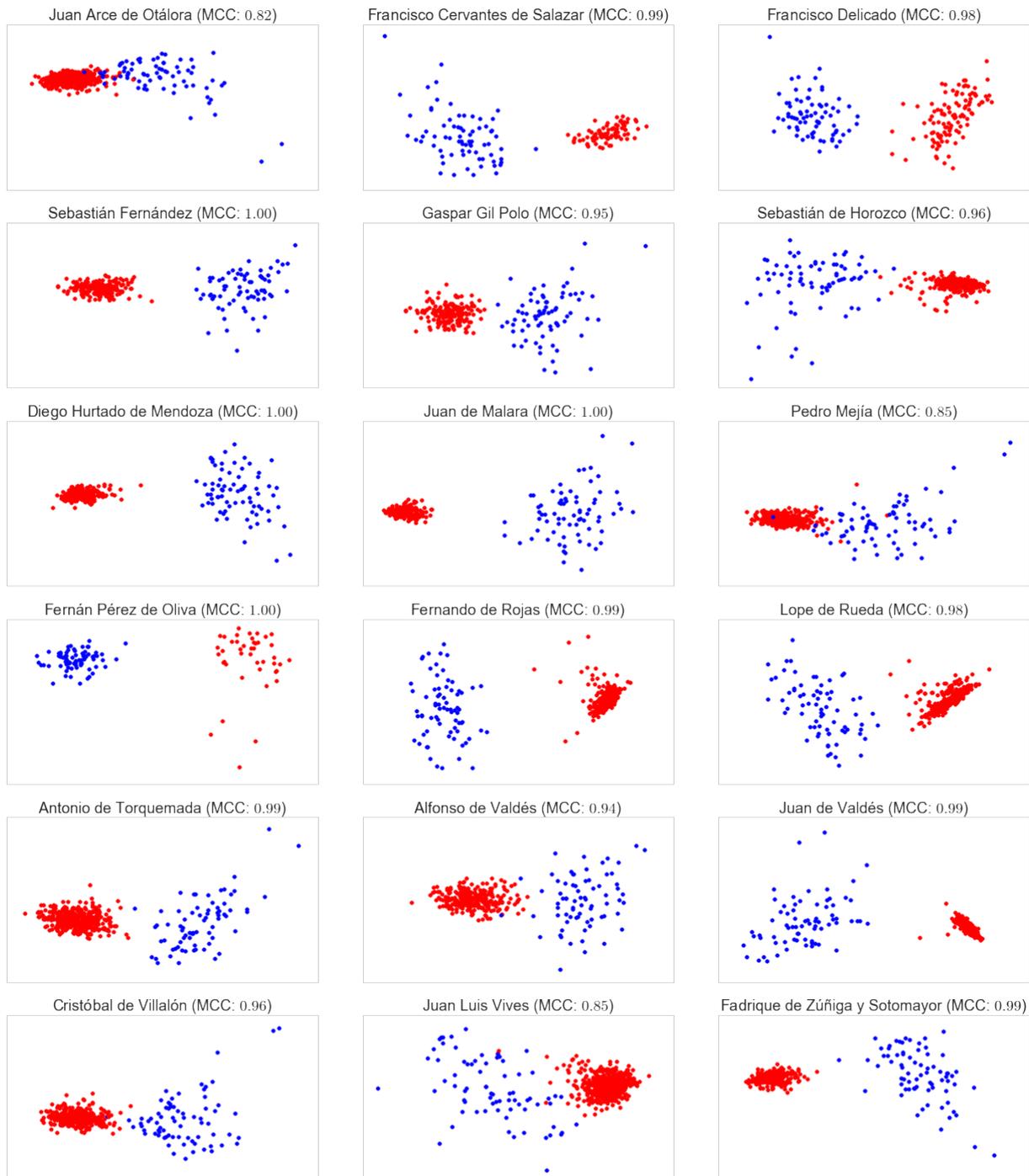

Figure 9: LDA of the 150 most common words in each pair from our corpus. Charts represent the 2 dimensions extracted by linear discriminant analysis of the frequency distribution of the 150 most common words in each pair of *Lazarillo* (blue) and the combined works of each of the possible candidates in the corpus (red). Only 600 random chunks of 300 words are represented, although all were taken into account during the analysis. Matthew's correlation coefficient (MCC) is added between parentheses.

Moreover, the naive feature set used by Burrows is not the only possible choice. Based on recent competitions for authorship attribution and author identification and verification (Argamon and Juola, Overview; Stamatatos et al., Overview), we extracted the features used by the winners (see table 4) and tested PCA under them for 2, 3, 5, 10, and 15 principal components, using the top 25, 50, 150, and 300 top words for vocabulary-based features. As the number of works per author is pretty limited in our corpus, making an instance-based analysis virtually impossible, we also segmented the texts in chunks of 300 words without breaking sentences, although only 600 chunks are represented in the charts for clarity reasons. After analyzing all possible combinations of this new setup in the search for a set of parameters that maximized the variance, a process usually known as grid search in the machine learning literature, we found that punctuation marks features, regardless of the number of words per chunk, provided the best result accounting for a variance of almost 48%. However, when plotted (see figure S1 in SM) there is no obvious way to separate the chunks of the *Lazarillo* and those belonging to the rest of authors. With the exception of perhaps Fernando de Rojas, the remainder turned out to be unusable in terms of identification of a possible author. Incrementing the number of components had a positive effect on the variance captured by the PCA, although we had to reach a balance between the number of principal components and the number of dimensions that can be represented in a chart and still be useful. By using 5 components we achieved a variance of around 80% with punctuation marks features, and after plotting the first 3 principal components, Sebastián Fernández, Diego Hurtado de Mendoza, Fernando de Rojas, and Lope de Rueda seem to be clustering separate from the *Lazarillo* (see figure S2 in SM). Higher number of PCs reported higher variance but were unsuitable for graphical representation.

Although revealing under certain circumstances, this capturing of the variance performed by PCA is not well suited for authorship attribution nor the general task of automatic classification ( Juola, «Authorship attribution» 233-334). As seen in our experiments, the dimensions that carry the most information does not necessarily have to be the ones that allow for an easier identification of the different clusters at game. An alternative technique that alleviates this limitation is the Linear Discriminant Analysis (LDA),[62] first formulated by Ronald A. Fisher in 1936 as a 2-class problem and later generalized for the multiclass scenario (Fisher, 179-188; Rao, 159-203). LDA is closely related to the analysis of variance (ANOVA, as applied by Holmes and Forsyth to the *Federalist*) and PCA, but in addition to finding the axes that maximize the variance, also finds the axes

---

62.– This analysis should not be confused with Latent Dirichlet Allocation (LDA), a technique formally presented in 2003 (although introduced in 2002) by David Blei, Andrew Ng, and Michael I. Jordan for topic modelling. It built on a series of improvements over previous techniques (specially from latent semantic analysis and its probabilistic version, LSI and pLSI), overcoming some of their limitations such as allowing its embedding into other methods. As proposed by its authors, LDA can be seen «as a dimensionality reduction technique, in the spirit of LSI, but with proper underlying generative probabilistic semantics that make sense for the type of data that it models,» specially when that data is a corpus of texts. The specific formulation of LDA is beyond the scope of this study, but generally it models each work from a corpus as a mixture of various topics, which is organized following a Dirichlet distribution. Since its conception, the technique has been successfully applied in a variety of subjects ranging from history to genomics. One recent area of application is precisely authorship attribution studies where LDA is usually combined with other methods and achieving good results. Unfortunately, we have not covered Latent Dirichlet Allocation in this study (Blei, Ng, and Jordan; Deerwester et al.; Papadimitriou et al.; Hofmann; Seroussi, Zukerman, and Bohnert; Savoy) that maximize the separation between different groups.[63] Recently LDA has been applied successfully to authorship attribution studies, although related literature on the topic is scarce (Stamatatos, Fakotakis, and Kokkinakis, 471-95). We tested LDA as a dimensionality reduction method under the same settings used for PCA. Our results show that LDA might convey more discriminatory power than PCA while capturing similar levels of variance. Figure 9 shows clearly separated clusters for all the authors but a few. This might suggest that the only viable candidacies might be Pedro Mejía, Alfonso de Valdés, Juan Arce de Otálora, and to a lesser extent Juan Luis Vives and Cristóbal de Villalón. Other discriminant methods exist: the family of *neighbors* methods take advantage of the visual representation, and group together elements based on the center of the cluster, the distance, and other parameters. A version known as nearest shrunken centroid has been reported to produce really good results in authorship attribution problems (Jockers, Witten, Criddle, 465-491; Schaalje et al., 71-88).

## The Rise of the Learning Machines

Variance and educated guesses over reduced dimensions of a feature set in a plot are generally accepted as (exploratory) authorship analysis and as indicators for further study. In order to properly assess their efficacy, we recur to the standard framework of machine learning measures. Machine learning is a subfield of computer science fed by pattern recognition, artificial intelligence, and computational statistics. At its core, it tries to construct algorithms that are able to learn from an input of known data (training samples) and make predictions or decisions on unseen data. Dependending on how much we know about the training samples we talk about supervised learning, if the samples are labeled as belonging to classes, or about unsupervised learning when those classes are still to be determined, either their number, membership, or both. PCA can be seen as an instance of unsupervised learning whereas LDA is supervised learning since it needs the labels of the training data to work. In fact, some machine learning methods are able to handle big feature sets without applying dimensionality reduction, *i.e.*, Support Vector Machines (SVM).[64] Regarding authorship attribution, a single training sample would be a text from a specific author, either in the instance- or profile-based mode, that is transformed into a numerical feature vector in a process of feature extraction; a labeled training sample would be the same text annotated with its author. In our context, classes would represent the authors of our corpus, and unsupervised methods try to find the clusters that better group the works of a same author together; while supervised learning methods learn by the examples in order to classify an anonymous into one or more of the classes. When categorized by the kind of output machine learning methods produce, classification and clustering are among the most relevant in authorship attribution studies. Other forms of machine learning include dimensionality reduction, that can also be of help, and regression or density estimation, specifically applied to continuum streams of data rather than discrete, as it is our case with authors.

---

63.– In doing so, it makes the assumption that the feature set (independent variables) is normally distributed

64.– For an introduction to the topic from the perspective of authorship attribution, we recommend Juola (*Authorship*).

Defining authorship attribution problems in the general context of machine learning allows us to apply its measures to the case under examination. Despite the existing debate around the authorship of some of the works considered in this study —*Diálogos* by Valdés—, we made the arguable initial assumption that our corpus only contains works of undisputed authorship, which places this study under the umbrella of supervised learning with a close-set corpus. The process goes as follows: first, training data (works labeled with their authors) is used to train or learn a model that can be binary if there are only two classes to decide, as in the work written by an author or the rest, or multiclass if the algorithm is able to deal with more than two, classifying each work to its author. Once the model is fitted with the training data a score is extracted to test the adequacy of the model. If the performance is good enough,[65] the model is asked to predict the label (author) of the unseen data (the *Lazarillo*). One way to assess this score is by holding out part of the training data and using it later in the prediction step for validation. Cross-validation might improve the results and reduce the problem of model overfitting —a model that predicts perfectly the training data but fails with new data— by randomly segmenting the training set several times (folds), fitting a model for each fold, and averaging the score following defined strategies that will be detailed later. Since held-out data is labeled (ground truth) we can calculate different measures based on the number of correct and wrong predictions. In classification tasks some commonly used measures are accuracy, precision, recall, and F-score ($F_1$). Given a class as an author and a data entry as a chunk of a work, accuracy is defined as the ratio of chunks correctly assigned to their true author (hits) divided by the total number of chunks by each author; precision as the ratio between one author hits divided by the total number of chunks correctly or incorrectly assigned to that author; recall or sensitivity as the ratio between one author hits and the total number of chunks existing by that same author; and F-score as the harmonic or weighted mean of precision and recall. In the context of binary classification, as it is the case in our LDA analysis, one class is considered «positive» and the other «negative,» leading to the definition of the measures in terms of true and false, positive and negative rates. These measures go from 0 to 1, where values closer to 1 are preferred. Another useful measure that arises in the 2-class problem is the Matthews correlation coefficient (MCC) that is well suited for tasks where the classes are of different sizes, as it is our case with only one work for the anonymous and sometimes several or lengthy works for the candidate authors (Powers). MCC is a measure of correlation that comprises true and false positive and negative rates (the confusion matrix), and it is considered one of the best measures for binary classification. Values of MCC range from -1, meaning a total disadjustment between ground truth and prediction, to +1, perfect prediction —a value of 0 would mean no better prediction power that a random prediction.[66] Figure 9 includes values of MCC between parenthesis for the binary classification task performed by LDA for each author against the author of the *Lazarillo*. In authorship verification problems, high precision is usually easier to achieve than high recall. In the LDA run and after a 10-fold cross-validation, Alfonso de Valdés accounted for the lowest values of recall (0.95) and accuracy (0.98), a result

---

65.– This actually depends on the field of study, the model, and the scoring method.

66.– It is sometimes compared to the area under the receiver operating characteristic (ROC) curve that results from plotting the true positive rate (recall) against the false positive rate, but it performs better with unbalanced classes.

that would suggest that none of the authors in our corpus would be the true author. In addition to the high values obtained for the lowest recall and accuracy, the Matthews correlation coefficient reported over 0.85 for all authors but Juan Arce de Otálora, Pedro Mejía, and Juan Luis Vives, which give us a quite dubious threshold to start considering them plausible candidates for the authorship. Furthermore, this method has been reported to perform poorly for authorship attribution even with models much more complex (Koppel, Schler, and Bonchek-Dokow, 1261-1276). The use of LDA as a discriminant by itself may produce misleading results since it might be affected by factors other than style. We must, therefore, further support such findings before making any hurried statement about a possible true author.

Fortunately, once we settle on using general machine learning approaches to authorship attribution, a whole range of possibilities opens up. Identifying the most likely author of the *Lazarillo* can be tackled from different angles. We can train a model for every pair of authors and assess the accuracy of the method by cross-validation. This approach is usually referred to in the literature as *one-versus-one*, as opposed to *one-versus-all*, where the models learn to distinguish an author against the rest. Classification then happens by a winner-takes-all strategy in one-versus-one, where the classifier with the best performance gets to decides the class—, and by a max-wins strategy in the one-versus-all case, in which each classifier adds a vote to a class based on its results, being the class with more votes the class that assigns the classification.

In order to test for the multiclass problem, we extracted the features defined in table 4 considering the whole corpus when vocabularies of words or characters were needed to be taken into account. Our first test using basic regression methods in a supervised fashion had very exciting results. We employed linear regression, Bayesian, and discriminant (neighbors) classification methods.[67] A profile-based version of the corpus was built with the texts segmented in chunks of at least 300 words without breaking paragraphs. Scores were averaged using a 10-fold cross-validation. Table 5 shows the 10 most performant algorithms (Ridge, Bernoulli, multinomial, and nearest centroid) and features sorted by their accuracy. Common n-grams, and bag-of-words are the features that report better results in our corpus, although our total feature set, a combination of all the features, behaves slightly better in every case. However, the increase in dimensionality that it involves might not be justified by the gain in precision, that barely adds up to a 0.12% in the worst case.

---

67.– Ridge classification is based on linear least squares; Bernoulli and Multinomial are specific cases of Naive Bayes classifiers; and nearest centroid can be related to discriminant analysis. Other classifiers tested with poorer results include Gaussian, Perceptron, k-nearest neighbors, radius neighbors, and nearest shrunken centroid.

Table 5: Top 10 algorithms and features pairs ranked by precision, recall and F-score.

| Algorithm | Features | Precision | Recall | F-score |
|-----------|----------|-----------|--------|---------|
| Ridge | total | 0.9718 | 0.9696 | 0.9701 |
| Ridge | cng | 0.9706 | 0.9675 | 0.9682 |
| Bernoulli | total | 0.9450 | 0.9273 | 0.9296 |
| Bernoulli | cng | 0.9429 | 0.9176 | 0.9215 |
| Multinomial | total | 0.9418 | 0.9273 | 0.9295 |
| Multinomial | cng | 0.9341 | 0.9078 | 0.9116 |
| Nearest centroid | cng | 0.9312 | 0.9067 | 0.9111 |
| Nearest centroid | total | 0.9211 | 0.9067 | 0.9092 |
| Bernoulli | bow | 0.9170 | 0.9078 | 0.9058 |
| Ridge | bow | 0.9287 | 0.8872 | 0.9032 |

In order to determine the most plausible author we used a max-wins strategy and also the average number of chunks assigned to each candidate. In this settings, Juan Arce de Otálora, who was assigned the most number of chunks most of the times, seems to be the winning author in both cases, with an important difference over the second ones in both the win and the average strategies, being those Gaspar Gil Polo and Alfonso de Valdés, respectively (see table 6). Interestingly, the result holds with or without *interpolaciones* and also for the second part of the little book —in which case Cristóbal de Villalón is also added. It is worth noting that the algorithm that reported the best performance grants the second position to Pedro Mejía instead of Gaspar Gil Polo. There seems to be an effect of the total number of chunks per author in the corpus over the predictions. The class imbalance problem is known to affect drastically the effectiveness of vector space models. Several approaches have been proposed in the last years to tackle this situation regarding authorship attribution (Stamatatos, «Author identification» 790-799). Segmenting or re-sampling the texts (reusing some parts of the text) in order to re-balance the number of samples per author is one of the methods proposed by Stamatatos for the instance-based approach. To alleviate the situation in the profile-based approach, we used a cut-off sampling approach by randomly removing the number of chunks that are over a fraction of the average number of chunks per author in the corpus, while resampling author texts whose number of chunks are said fraction below the average —we used a chunk fraction of 10%. We then averaged results over several general machine learning methods using 10-fold cross-validation.

Table 6: Top authors with the most chunks of the Lazarillo assigned to them for the different methods and features. Number of pairs algorithm and feature set wins, and the average number of chunks assigned for each author are included in the last two columns.

|  | Ridge | | | Bernoulli | | | Multinomial | | N. Centroid | | Wins | Avg. |
|---|---|---|---|---|---|---|---|---|---|---|---|---|
|  | total | cng | bow | total | cng | bow | total | cng | cng | total | | |
| JAO | 34 | 47 | 37 | 42 | 54 | 18 | 48 | 60 | 46 | 39 | 9 | 42.50 |
| AV | 7 | 3 | 7 | 12 | 6 | 18 | 11 | 5 | 5 | 10 | 0 | 8.40 |
| GGP | 2 | 1 | 4 | 10 | 2 | 21 | 12 | 4 | 9 | 14 | 1 | 7.90 |
| PM | 22 | 17 | 16 | 0 | 0 | 1 | 0 | 0 | 2 | 1 | 0 | 5.9 |
| LR | 2 | 3 | 1 | 8 | 11 | 6 | 1 | 4 | 6 | 1 | 0 | 4.30 |
| CV | 1 | 0 | 3 | 1 | 0 | 7 | 1 | 0 | 5 | 8 | 0 | 2.60 |

Support Vector Machines (SVMs) are binary classifiers in nature and as such they recur to ensemble techniques to generalize to the multiclass version. They are intended to work with high-order feature vectors by finding a hyperplane (vector) that allows (supports) the division of the feature space in two spaces, while maximizing the average of the distances from the features vectors to such hyperplane. In a way, they automatize the visual inspecting task we performed for exploring the results of the LDA. Used in combination with bag-of-words or character n-grams, SVMs are a solid choice for authorship attribution, from newspaper articles, to e-mails or 19th-century English literature. Their most important characteristic, and the reason why they became so popular, is that they can handle several thousands of features without resulting in overfitting or needing preprocessing steps (Teng et al., 1204-1207; Sanderson and Guenter, 482-491; Joachims). Other models that have reported good results in authorship attribution problems include neural networks, decision trees, maximum entropy, memory-based learners, and ensemble learning methods.[68] Faced with the impossibility of testing every single existing method, we resorted once again to the winners of several authorship attribution competitions editions that included Spanish corpora, and when suitable, according to the specifics of our corpus, we tested some of the best performing methods with the feature sets we defined in table 4 (Argamon and Juola, Overview; Juola, «An Overview»; Rangel et al; Stamatatos et al., «Overview»; Stamatatos et al., «PAN 2015»). Specifically, we tested linear and nonlinear SVMs;[69] maximum entropy learning (MaxEnt), a type of logistic regression method (not to be confused with linear regression) that measures the relationship between features and their assigned author using a logistic function for estimating the probabilities (Nigam, Laf-

68.– Especially promising is the application of biologically inspired neural networks, such as recurrent and convolutional neural nets, that have reported results that outperform state-of-art for the Spanish case (Bagnall).

69.– Nonlinear SVMs use transformations of the feature space, specifically we used a gaussian kernel (RBF). For an introduction to kernel-based methods in machine learning in general we recommend Nello Cristianini and John Shawe-Taylor.

ferty, and McCallum, 61-67); and random forests, an ensemble technique that reduces the overfitting problem in decision trees by building a number of them and classifying unseen samples as the most repeatedly assigned label (the statistical mode) (Maitra, Ghosh, and Das; Pacheco, Fernandes, and Porco). We also included other less performant algorithms that showed some good results for the Spanish case: stochastic gradient descent classification (SGD), an optimization-based method that can operate with large datasets since only takes one sample at a time, and although it might not find the optimum, most of the times it finds a reasonably good approximation (Caurcel Díaz and Gómez Hidalgo); and bagging, an ensemble classifier that trains decision trees, although other learners can be used, on random subsets of the features and combines their prediction by voting (Giraud and Artières). When suitable we normalized the feature vectors and reduced their dimensionality up to 100 components prior cross-validation.[70] Table 7 shows the 10 best performing algorithms with their respective feature sets. They all performed extremely well, especially maximum entropy and linear SVMs, and the only difference is the feature set: common characters 3-grams and our total fusion of features are again dominating.

Table 7: Top 10 supervised algorithms and features pairs ranked by precision, recall and F-score without using dimensionality reduction

| Algorithm | Features | Precision | Recall | F-score |
| --- | --- | --- | --- | --- |
| Max Ent | total | 0.9762 | 0.9740 | 0.9745 |
| Max Ent | cng | 0.9723 | 0.9707 | 0.9712 |
| Linear SVM | total | 0.9700 | 0.9685 | 0.9689 |
| Linear SVM | cng | 0.9682 | 0.9664 | 0.9668 |
| SVM | cng | 0.9558 | 0.9458 | 0.9480 |
| SVM | total | 0.9563 | 0.9447 | 0.9474 |
| SGD | total | 0.9512 | 0.9382 | 0.9406 |
| Max Ent | bow | 0.9438 | 0.9382 | 0.9397 |
| SGD | bow | 0.9375 | 0.9273 | 0.9302 |
| SGD | cng | 0.9430 | 0.9262 | 0.9000 |

We then used the most performant models to classify the chunks of the *Lazarillo* to one of the candidates (see table 8) finding that Juan Arce de Otálora beat the rest of the authors in both the max-wins —9 over 1— and the average criteria —almost 37 out of the 73 chunks of the *Lazarillo* are always assigned to Otálora regardless of the method.[71] Second positions correspond to Alfonso de Valdés in max-wins and Pedro de Mejía in chunk average. This results strongly points out at solid similarities between the writing style of the little book and the work by Juan Arce de Otálora. We believe that despite the limitations in our corpus and the candidates chosen to represent the debate around the

70.– In fact, we tested with and without dimensionality reduction, and with PCA and LDA, and even after the fact that supervised decomposition as the one performed by LDA might bias cross-validation, with obtained very similar results and a general speedup when applied.

71.– Results hold with or without interpolations, although for the second part of the little book Cristóbal de Villalón seems to be slightly stronger than Otálora.

possible author, an average of half the chunks assigned to Otálora —ranging from 33% under a SGD learner with precision of 94% and bag-of-words features to more than 86% of the chunks under a nonlinear SVM with precision of 96% using common 3-grams—, is a strong and data-based argument in favour of the candidacy of the jurist.

Table 8: Top authors with the most chunks of the Lazarillo assigned to them for the different methods and features. Number of pairs algorithm and feature set wins, and the average number of chunks assigned to each author are included in the last two columns.

| | MaxEnt | | | LinearSVM | | SVM | | SGD | | | Wins | Avg. |
|---|---|---|---|---|---|---|---|---|---|---|---|---|
| | total | cng | bow | total | cng | cng | total | total | bow | cng | | |
| JAO | 39 | 42 | 15 | 34 | 40 | 63 | 58 | 37 | 24 | 44 | 9 | 36.82 |
| PM | 14 | 14 | 7 | 16 | 15 | 9 | 8 | 12 | 14 | 12 | 0 | 11.00 |
| AV | 13 | 8 | 25 | 13 | 6 | 0 | 4 | 4 | 10 | 0 | 1 | 7.64 |
| GGP | 4 | 5 | 17 | 5 | 6 | 0 | 0 | 0 | 19 | 0 | 0 | 5.09 |
| JLV | 0 | 1 | 0 | 0 | 0 | 0 | 1 | 20 | 3 | 17 | 0 | 3.82 |
| CV | 1 | 1 | 5 | 1 | 3 | 0 | 0 | 0 | 2 | 0 | 0 | 1.18 |
| LR | 1 | 2 | 1 | 1 | 2 | 1 | 2 | 0 | 1 | 0 | 0 | 1.00 |

## Unmasking the Author of the *Lazarillo*

At the beginning of our study we did our best effort to collect a set of works that would sufficiently represent the stronger candidates in the debate about the authorship of the *Lazarillo*. The reason behind was to allow the use of statistical methods in order to analyze the problem as a closed-set task. However, our best set of classifiers, even when not overfitting, would always assign chunks of any given book to the authors that have been trained on. That is the fundamental flaw of the closed vs open-set problem. We believe that a consistent prediction of more than half the chunks to Juan Arce de Otálora is not casual, but when asked with the task of classifying an unseen work, the regular supervised methods we employed lack a foundation to decide «none of the above» as the right answer. In 2004 Moshe Koppel and Jonathan Schler proposed —and improved in successive years— a new ensemble method to tackle this issue (Koppel and Schler, «*Authorship verification*»; Koppel, Schler, and Argamon, *Authorship*; Koppel, Schler, and Bonchek-Dokow, *Measuring*). We used their method, based on feature elimination,[72] in an attempt to dispel the last doubts about the author of the *Lazarillo*, considering now the problem of its authorship as open-set.

Ensemble learning techniques usually provide better results and predictive power than their algorithms would separately. Koppel and Schler *unmasking* method is one of the best-known techniques of its kind, albeit having numerous subtleties that need to be fine tuned corpus-wise. A defining characteristic of their technique is the ability to decide not only whether an anonymous text is written by one of the authors in the candidate set, but also if the text has not been written by any of them. In its general form it conceives the au-

---

72.– It has been noted and we agree on certain similarities between the unmasking method and a technique known as feature elimination used in cancer classification (Guyon et al., 389-422; Huang and Kecman, 185-194).

thorship problem as a one-class classification task built upon linear SVMs. Although the specifics of its implementation, which we had to develop in Python in the lack of reference source code, are out of the scope of this study, the main idea remains rather intuitive. Given a set of features for a pair of works the method iteratively removes «those features that are most useful for distinguishing between [them]» and «gauge the speed with which cross-validation accuracy degrades as more features are removed.» Koppel and Schler hypothesize that if two works are written by the same author then «whatever differences there are between them will be reflected in only a relatively small number of features, despite possible differences in theme, genre and the like.»[73] For each pair <work, candidate's works> in the corpus,[74] a linear SVM is built to distinguish between them. The feature set is bag-of-words-like, with the $n$ most frequent words calculated as the average of the frequency in the work and the candidate's works for a given pair. In a number of steps $m$, the top $k$ most and least informative features are removed and the accuracy of the SVM is measured using a 10-fold cross-validation. These $n$ values of accuracy that define the degradation curve are used to build a vector of «essential features» that is labeled *same-author* if the work was in fact written by the candidate author in the pair, and *different-author* otherwise. Figure 10 shows an example of degradation curves for the work *Las Disciplinas* by Juan Luis Vives against the rest of the candidate authors with default parameters as defined by Koppel and Schler ($n$=250, $k$=6, $m$=8). The method assumes that these two types of curves are different and easy to identify. A linear SVM is then trained to distinguish between *same-author* and *different-author* curves. When asked to decide on an unseen work, degradation curves are built for each of the candidate authors in the corpus, and then the SVM decides if any of the unseen work degradation curves are classified as *same-author*, and in that case return for which one. The method does not guarantee that an author will be returned and it does not prevent more than one author from being the result. Using Matthew's correlation coefficient, we obtained a classification score 0.98.

Due to the computationally expensive nature of the method, it is usually a good idea to reduce the number of authors and works in the corpus, although it is proven that the unmasking behaves better with lengthy texts such as books (Sanderson and Guenter, *Short text*). Building upon our previous results, we can now shrink the pool of candidates to those that have shown to be likely authors in the previous methods along this study. Nonetheless, it is worth noting that some of the candidates that we thought to be mere impostors are now among the most plausible ones, *i.e.*, Pedro Mejía. We must interpret this as part of the *Lazarillo* sharing stylistic similarities with the works of others, and consequently when reducing our pool of candidates to reduce execution time, we must get rid of those authors who were assigned in average less than one chunk of the little book. The final list of authors considered for unmasking includes Juan Arce de Otálora, Pedro Mejía, Alfonso de Valdés, Gaspar Gil Polo, Lope de Rueda, and Juan Luis Vives. Just an ironic coincidence that, as the Avellaneda's song goes, 6 can be the most likely authors of the little book. Moreover, we must highlight the recurrent apparition in our analysis of Cristóbal de Villalón, not only among the possible authors but as the most assigned author of the second part of the *Lazarillo*; thus

73.– The efficacy of the method in a cross-genre setup was later confirmed by Mike Kestemont et al. (340-356).

74.– If for a certain pair, the work in question is by the candidate, we remove said work from the candidate's works for that pair.

we included him as well. We calculated all the curves and essential feature vectors for the *Lazarillo* against the candidates in our corpus, and the trend shown in figure 11 seems to confirm that Juan Arce de Otálora shares the most stylistic similarities with the little book, followed closely once again by Alfonso de Valdés, as their drop in accuracy per iteration is larger than for the rest of the authors. Unfortunately, we cannot state with enough certainty that either Arce de Otálora or Valdés is the true author, since the SVM that distinguished between *same-* and *different-author* curves did not assign a clear winner; it returned *different-author* for all the authors. Nevertheless, this last result is the last of a series of methods applied along this study that support Juan Arce de Otálora as the most likely author. The result, however, demands more fine tuning of the parameters of the unmasking method.

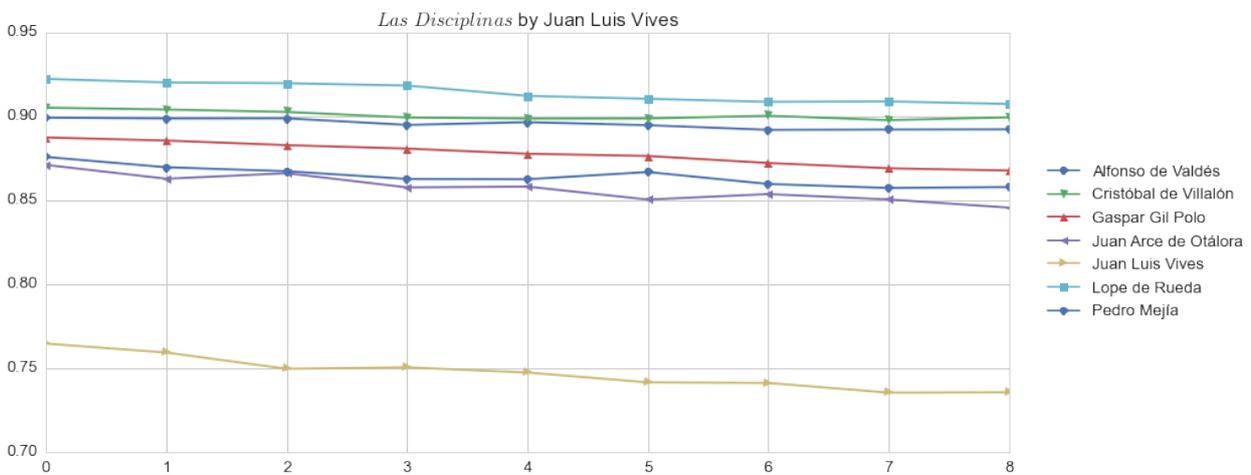

Figure 10: Unmasking *Las Disciplinas* by Juan Luis Vives against each of 6 authors (*n*=250, *k*=3). The curve below all the authors is that of Juan Luis Vives, the actual author.

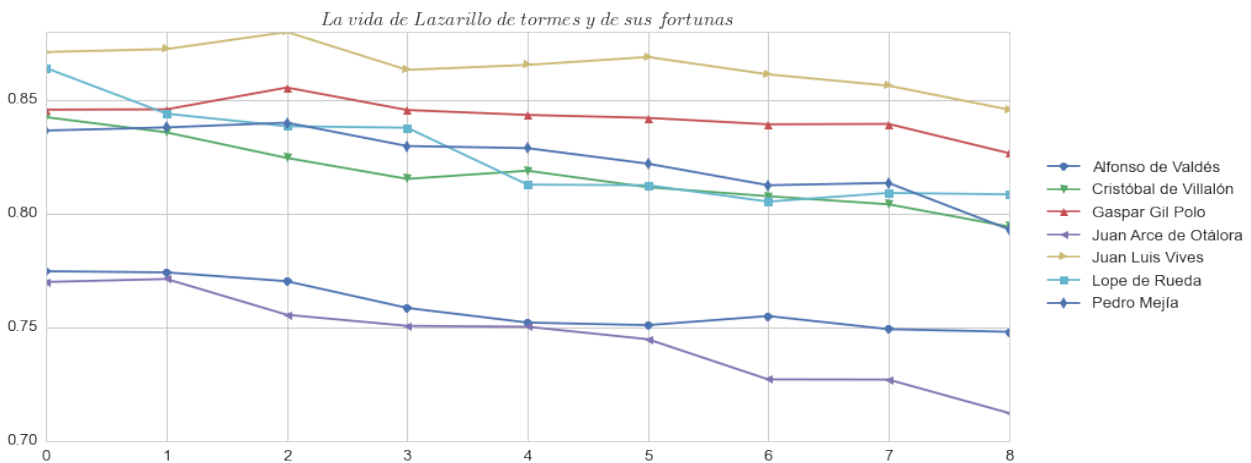

Figure 11: Unmasking *Lazarillo* against each of 6 authors (*n*=250, *k*=3). The curve below all the authors is that of Juan Arce de Otálora, the most likely author, followed by that of Alfonso de Valdés.

## Discussion

Coinciding with the statistical approach carried out by Madrigal, Juan Arce Otálora has been consistently assigned high positions in the analysis of the authorship of the *Lazarillo*, but if we are to accept the result by Burrows's Delta and Koppel and Schler unmasking method, the evidence is not enough to support him as being the true author: both methods agree on the prominent similarity between Arce de Otálora and Alfonso de Valdés' writing styles, but suggest that any of them is in fact the author. We add, nevertheless, that the candidacy for Arce de Otálora has been strongly supported. Deficiencies are in general attributable to the corpus rather than the methodology. Diego Hurtado de Mendoza, on the other hand, one the most documented candidates of all and possibly the one towards whom we felt more confident, turned out not to be a strong player in our analysis. We believe that one of the reasons is the lack of representation of his works in our corpus. Hurtado's *De la Guerra de Granada* might not be the best work to put on play against *Lazarillo*, or at least not the only one, as the linguistic registry is very different in both cases. As Ángel González Palencia pointed out, Hurtado de Mendoza's informal style expressed in his personal letters would account for a better representation in the corpus, but he also considers that style of writing not to be precisely descriptive of Mendoza's, as letters had to be usually written with haste. That is, if the style of *Lazarillo* were to be similar to Mendoza's, it would have to be similar to a style which he never would use to write prose, unless it was written as a joke for a then young prince, as Agulló argues.

In either case, as the study advanced, we tried to minimize the effect of the class imbalance problem, and when critical for certain methods, it turned out not to be such an obstacle. Alfonso de Valdés, whose works were not precisely the longest ones, still had been consistently given as one likely author. This study might sustain Valdés' candidacy in relation to the internal evidence when compared to the little book, oftentimes the reason of the criticism to Navarro Durán's candidate.

Juan Luis Vives, the candidate with the longest corpus and brought to the discussion in the initial exploration methods, was ultimately not sufficiently supported by any of the supervised learning techniques. A similar case is Lope de Rueda, who showed in the methods affected by the imbalanced-class problem but disappeared later. Other authors such a Fernán Pérez de Oliva or Fernando Delicado were soon removed from the debate. For those that were not part of the impostors we believe that this study is proof enough to reject their candidacies. And for the impostors that in the end resulted to share stylistic similarities with the little book, we believe there is a demand for further research in their cases, as for Pedro Mejía as a possible contributor of the *Lazarillo*, or even Cristóbal de Villalón for the second part, which deserves its own study. The hypothesis of a multiple authorship might also be backed up if we only consider the style markers evidence brought up by our study, and although not accountable or usable by literary critics as the features sets that carried the most discriminative power were undecipherable in a human context, we provide with stylistic proof that might support the idea.

## Conclusions

This study started with an overview of the status of the question of the authorship of the *Lazarillo*, which allowed us to establish a baseline corpus of candidates to work with. The subsequent exploratory analysis employing distance-based measures and methods from unsupervised learning started to give the first hints. Juan Arce de Otálora and Alfonso de Valdés were then highlighted and soon supported by the use of more sophisticated methods. The majority of the statistical evidence seem to point out in the direction of Arce de Otálora by a wider margin with regards to Valdés, and while our corpus is not as comprehensive as the one used by Madrigal, the jurist is still chosen by the learning methods as the most likely author. It seems as if all statistical techniques agree on Arce de Otálora, which supports the hypothesis of Madrigal, but it might not be the ultimate proof the authorship needs. Open-set methods suggest that none of the authors wrote the little book. After all, if, as Francisco Rico mentions in his 2011 edition (Anónimo ed. Rico, 128), the *Lazarillo* was the only work written by his author, any method, computational or not, based on the comparison of styles, mentions, idioms, or fingerprints, turns out to be useless. Under such assumption and due to the lack of other texts used as clues, the traditional historiographic profile-based research stands out as our only chance to find the author.

The Erasmian answer to the question of the authorship is recursively based on the principle of authority: it is important to unmask the anonymous of a work if the writer is in fact an important author. In recent times, the author might not ever be of interest at all, as the Barthesian conception of the death of the author considers. Others, however, agree on that knowing the author of a work «changes its meaning by changing its context [...,] certain kinds of meaning are conferred by its membership and position in the book or oeuvre» (Love, *Attributing*, 46). Paraphrasing Love, *Lazarillo* by Diego Hurtado de Mendoza, with its life parallels and allusions, is a different story that *Lazarillo* by Friar Juan de Ortega or Pedro Mejía. While this study helps to dispel doubts around some of the most often cited authors for the little book, we still believe that the authorship of the *Lazarillo* plays an important role in the work. Unlike Américo Castro, we do not give much importance to the fact of the anonymity itself but to the actual 400-year-old debate about who the author might be. Discovering new authors and arguing in favor or against them injects with life the adventures of such Lázaro de Tormes. Every time a new author is proposed, a new reading is found in the *Lazarillo*. Because of this, part of us hopes nobody ever finds the definitive factual proof to prove the authorship, as that would take away all the fun from it.

## Further Research

Much is still to be done regarding computational approaches for the resolution of the anonymity of the *Lazarillo*. Forensic linguistics also includes problems related to author plagiarism and author clustering, which could help to identify, for example, the legitimacy of the *interpolaciones* as part of the text of the Lazarillo, or to discern whether different hands intervened in the creation of the little book. Debates in this context, however, can also be enriched by the use of modern techniques such as those of the social network analysis. Previous studies in different areas have proven to be useful in shedding some light and contributing to the discussion of similar questions by the study of the graph structure of the actors involved (Suárez, Sancho, and de la Rosa, 281-285; Suárez et al., fqt050; Suárez, Sancho Caparrini, and de la Rosa; Suárez, McArthur, and Soto-Corominas, 45-50). While the use of this technique for authorship attribution would hardly result in a final answer, it configures an interesting path worth exploring in further research.

Compiling a better corpus to test authorship verification for each of the authors is another important future direction for investigation. Adding more authors and more works to the corpus could only benefit the study of the authorship of the little book. If both individual and institutional efforts were to be combined, the anonymity of the *Lazarillo* could be solved once and for all. Hundreds of mathematicians were able to altruistically combine their efforts to solve century-old problems (Gowers and Nielsen, 879-81; Cranshaw and Kittur, 1865-1874), therefore we believe that literary experts could do so as well for the *Lazarillo*. Having access to the digital editions that presumably RAE's CORDE handles as its core, or agreements with the editors of critical editions of Spanish Golden Age literature in order to use the same normalization rules for the old Spanish language, are only a couple of suggestions that could skyrocket the research on the topic. Moreover, proper coordination and agile communication channels to share early discoveries would be key factors to take into account. Traditional and nontraditional studies need to handshake and start a path together if we aim to find that elusive author of the masterpiece that is *The Life of Lazarillo de Tormes and of His Fortunes and Adversities*.

*Acknowledgments.* We acknowledge the support of the Social Sciences and Humanities Research Council of Canada through a Major Collaborative Research Initiative and the Canada Foundation for Innovation through the Leaders Opportunity Fund.

## Supplementary Materials

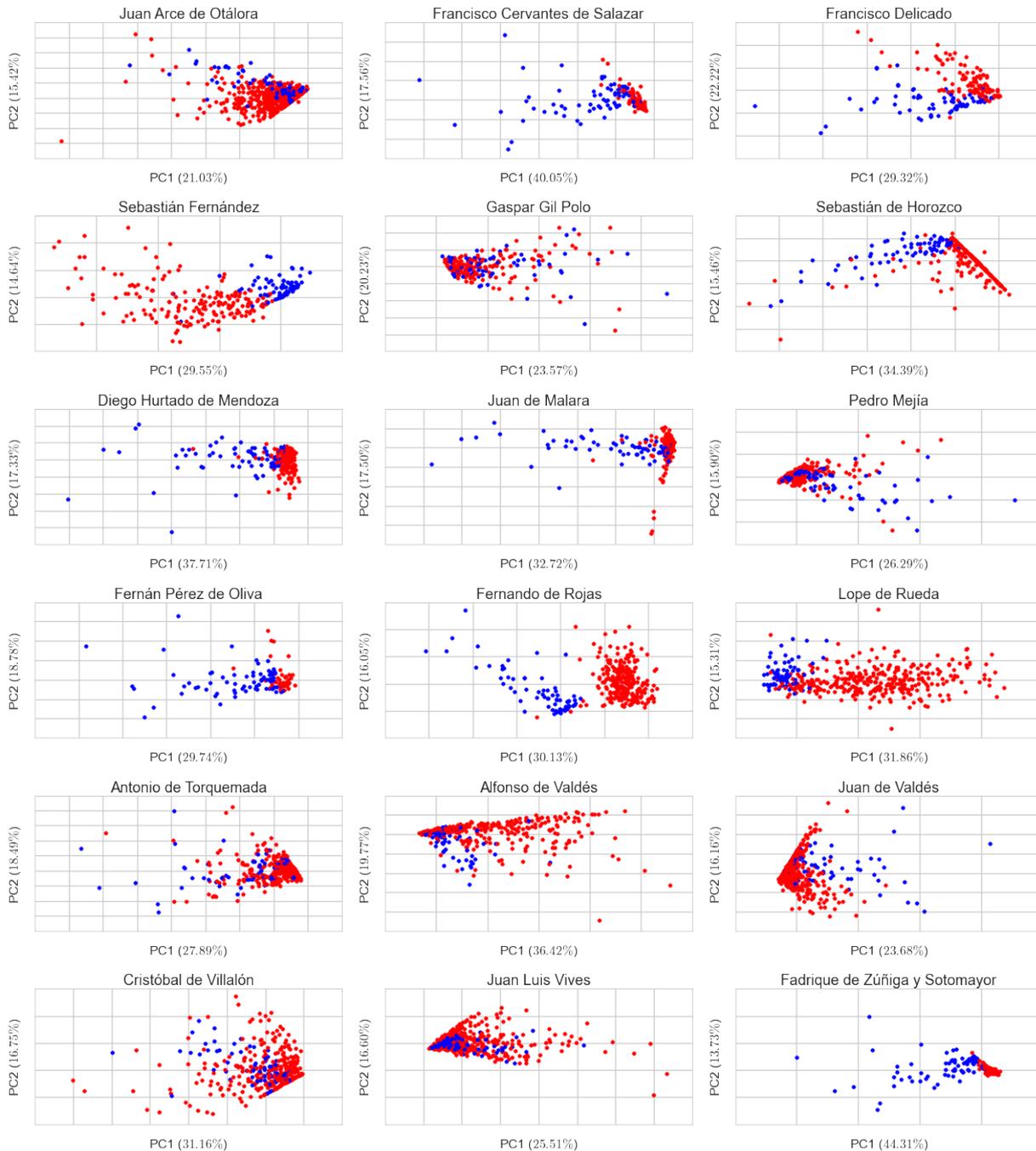

Figure S1: PCA of punctuation marks in our corpus. Charts represent the 2 principal components vectors of the frequency distribution all Spanish punctuation marks in the *Lazarillo* (blue) and the combined works of each of the possible candidates in the corpus (red). Only 600 random chunks of 300 words are represented, although all were taken into account during the analysis. Variance is shown as axes labels.

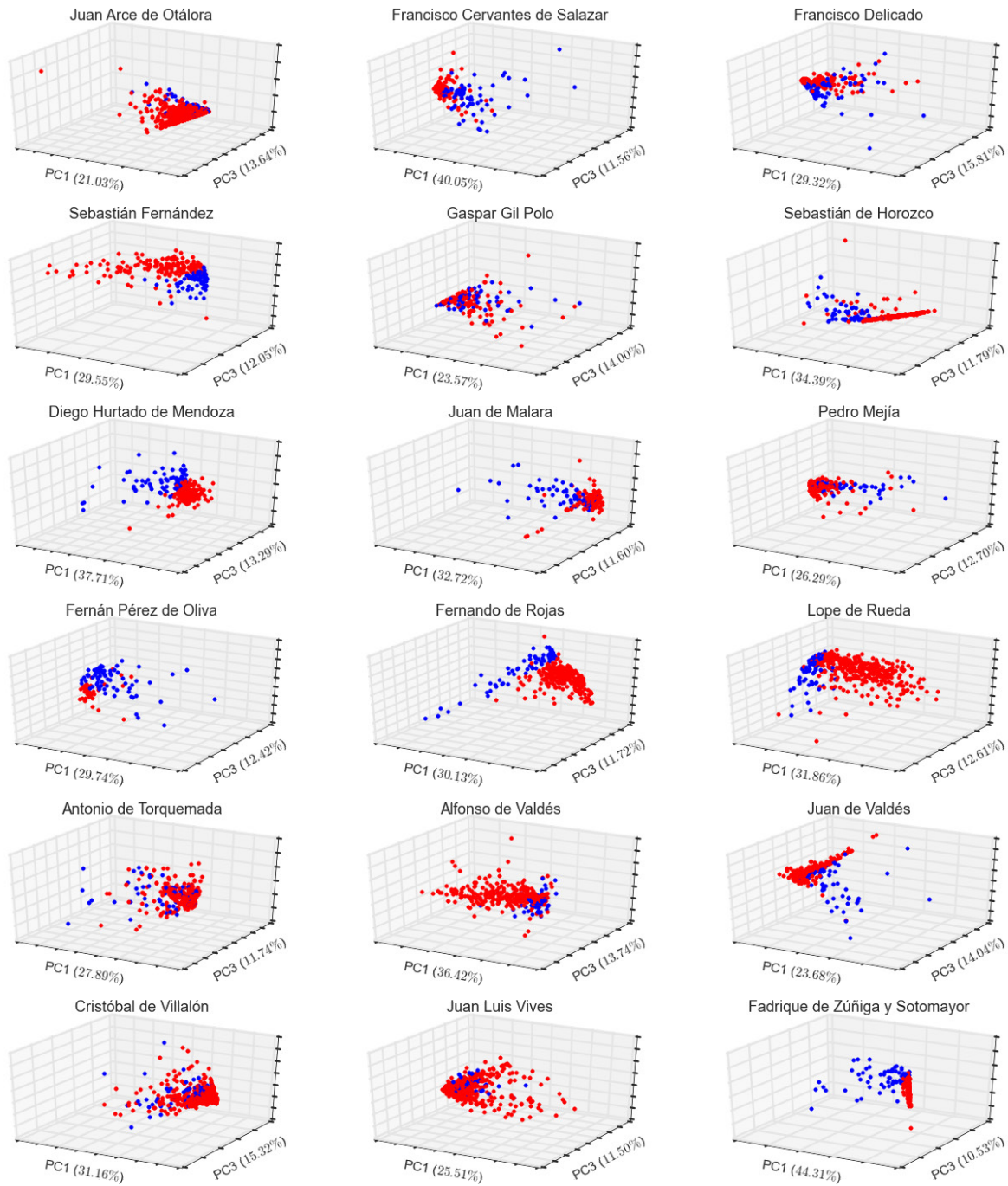

Figure S2: PCA of punctuation marks in our corpus. Charts represent the first 3 principal components of a 5 PCA of all Spanish punctuation marks in the *Lazarillo* (blue) and the combined works of each of the possible candidates in the corpus (red). Only 600 random chunks of 300 words are represented, although all were taken into account during the analysis. Variance is shown as axes labels.

Table S1: Timetable of attributions. Chronology of the candidates for the authorship of the *Lazarillo*, their support and their criticism. A dagger (†) besides the name of a possible author refers to him being proposed for the first time.

| Year | Author | Supported by | Criticized by |
|------|--------|--------------|---------------|
| 1605 | Juan de Ortega† | José de Sigüenza | |
| 1607 | Diego Hurtado de Mendoza† | Valerio Andrés Taxandro | |
| 1608 | Diego Hurtado de Mendoza | Andrés Schott | |
| 1624 | Juan de Ortega | | Tomás Tamayo de Vargas |
|      | Diego Hurtado de Mendoza | Tomás Tamayo de Vargas | |
| 1867 | Sebastián de Horozco† | José María Asensio | |
| 1873 | Diego Hurtado de Mendoza | Nicolás Antonio | |
| 1888 | Diego Hurtado de Mendoza | | Alfred Morel-Fatio |
|      | Juan de Valdés† | Alfred Morel-Fatio | |
| 1901 | Lope de Rueda† | Fonger de Haan | |
| 1914 | Juan de Valdés | | Julio Cejador y Frauca |
|      | Lope de Rueda | | Julio Cejador y Frauca |
|      | Sebastián de Horozco[75] | Julio Cejador y Frauca | |
| 1915 | Sebastián de Horozco | | Emilio Cotarelo |
| 1943 | Diego Hurtado de Mendoza | Ángel González Palencia | |
|      | Diego Hurtado de Mendoza | Eugenio Mele | |
| 1954 | Juan de Ortega | Marcel Bataillon | |
| 1955 | Pedro de Rhúa† | Arturo Marasso | |
| 1957 | Sebastián de Horozco | Francisco Márquez Villanueva | |
| 1959 | Juan de Valdés | Manuel J. Asensio | |
|      | Juan de Valdés | | Erika Spivakovsky |
| 1960 | Juan de Valdés | Manuel J. Asensio | |
| 1961 | Diego Hurtado de Mendoza | Erika Spivakovsky | |
| 1963 | Diego Hurtado de Mendoza | Olivia Crouch | |
| 1964 | Hernán Núñez de Toledo† | Aristides Rumeu | |
|      | Lope de Rueda | Fred Abrams | |
| 1966 | Juan de Ortega | Claudio Guillén | |
| 1969 | Diego Hurtado de Mendoza | Charles Vincent Aubrun | |
| 1970 | Diego Hurtado de Mendoza | Erika Spivakovsky | |
| 1973 | Sebastián de Horozco | José Gómez-Menor Fuentes | |

75.– Although José María Asensio was the first to suggest Sebastián de Horozco, the attribution owns much more to Julio Cejador y Frauca.

| | | | |
|---|---|---|---|
| 1976 | Alfonso de Valdés[†] | Joseph V. Ricapito | |
| 1978 | Sebastián de Horozco | Jaime Sánchez Romeralo | |
| 1980 | Lope de Rueda | Jaime Sánchez Romeralo | |
| | Sebastián de Horozco | Fernando González Ollé | |
| 1987 | Lope de Rueda | | Francisco Rico |
| | Sebastián de Horozco | | Francisco Rico |
| | Hernán Núñez de Toledo | | Francisco Rico |
| 1988 | Juan de Ortega | Claudio Guillén | |
| 1992 | Juan de Valdés | Manuel J. Asensio | |
| 2002 | Alfonso de Valdés | Rosa Navarro Durán | |
| | Alfonso de Valdés | | Antonio Alatorre |
| | Juan de Ortega | Antonio Alatorre | |
| 2003 | Lope de Rueda | Alfredo Baras Escolá | |
| | Alfonso de Valdés | Rosa Navarro Durán | |
| | Alfonso de Valdés | Juan Goytisolo | |
| | Francisco Cervantes de Salazar[†] | José Luis Madrigal | |
| | Alfonso de Valdés | | Antonio Alatorre |
| | Alfonso de Valdés | | Félix Carrasco |
| 2004 | Alfonso de Valdés | | Félix Carrasco |
| | Alfonso de Valdés | | F. Márquez Villanueva |
| | Alfonso de Valdés | | Valentín Pérez Venzalá |
| 2006 | Alfonso de Valdés | Rosa Navarro Durán | |
| | Alfonso de Valdés | | M. Antonio Ramírez López |
| | Alfonso de Valdés | | Francisco Calero |
| | Lope de Rueda | | Francisco Calero |
| | Juan Luis Vives[†] | Francisco Calero | |
| 2007 | Alfonso de Valdés | | Pablo Martín Baños |
| 2008 | Pedro de Rhúa | Francisco Calero[76] | |
| | Francisco Cervantes de Salazar | | José Luis Madrigal |
| | Juan Arce de Otálora† | José Luis Madrigal | |
| 2010 | Diego Hurtado de Mendoza | Mercedes Agulló | |
| | Diego Hurtado de Mendoza | Jauralde Pou | |
| | Diego Hurtado de Mendoza | | Javier Blasco |
| | Alfonso de Valdés | Rosa Navarro Durán | |
| | Alfonso de Valdés | Joseph V. Ricapito | |
| | Diego Hurtado de Mendoza | | José Luis Madrigal |
| | Diego Hurtado de Mendoza | | Rodríguez Mansilla |
| | Juan Arce de Otálora | Rodríguez López-Vázquez | |
| | Juan Arce de Otálora | | Rodríguez López-Vázquez |
| | Juan de Pineda† | Rodríguez López-Vázquez | |
| 2011 | Juan Arce de Otálora | | Francisco Calero |
| | Diego Hurtado de Mendoza | Mercedes Agulló | |
| | Diego Hurtado de Mendoza | Reyes Coll-Tellechea | |
| 2012 | Juan Luis Vives | M. Antonio Coronel Ramos | |
| 2014 | Juan Luis Vives | | Encarna Podadera |
| | Diego Hurtado de Mendoza | Joaquín Corencia Cruz | |
| | Juan Arce de Otálora | José Luis Madrigal | |

76.– As explained before, Francisco Calero suggests that Pedro de Rhúa and Juan Luis Vives were in fact the same person.